\documentclass[10pt,journal,compsoc]{IEEEtran}
\usepackage{amsmath,amsfonts}
\usepackage{array}
\usepackage[caption=false,font=normalsize,labelfont=sf,textfont=sf]{subfig}
\usepackage{textcomp}
\usepackage{color}
\usepackage{stfloats}
\usepackage{url}
\usepackage{verbatim}
\usepackage{graphicx}
\usepackage{cite}
\usepackage{soul}
\usepackage{enumerate}
\usepackage{enumitem}
\usepackage{xcolor}

\usepackage{amsmath,amssymb,amsfonts}
\usepackage{textcomp}
\usepackage{multirow}
\usepackage{booktabs}
\usepackage{graphicx}
\usepackage{textcomp}
\usepackage{booktabs}
\usepackage[hidelinks]{hyperref}
\usepackage{colortbl}
\usepackage{url}

\usepackage[linesnumbered,ruled]{algorithm2e}
\makeatletter
\newcommand{\removelatexerror}{\let\@latex@error\@gobble}
\makeatother

\newtheorem{problem}{Problem}
\newtheorem{definition}{Definition}

\begin{document}

\title{Multi-scale Traffic Pattern Bank for Cross-city Few-shot Traffic Forecasting}

\author{Zhanyu Liu,
Guanjie Zheng
\thanks{Zhanyu Liu, Guanjie Zheng are with Shanghai Jiao Tong University, Shanghai 200240, China. (Email: \{zhyliu00, gjzheng\}@sjtu.edu.cn)},
Yanwei Yu
\thanks{Yanwei Yu is with the Ocean University of China, Qingdao 266100, China. (email: yuyanwei@ouc.edu.cn)}
\thanks{Corresponding Author: Guanjie Zheng, gjzheng@sjtu.edu.cn}

}

\markboth{Journal of \LaTeX\ Class Files,~Vol.~14, No.~8, August~2021}%
{Shell \MakeLowercase{\textit{et al.}}: A Sample Article Using IEEEtran.cls for IEEE Journals}


\IEEEtitleabstractindextext{%
\begin{abstract}
Traffic forecasting is crucial for intelligent transportation systems (ITS), aiding in efficient resource allocation and effective traffic control. 
However, its effectiveness often relies heavily on abundant traffic data, while many cities lack sufficient data due to limited device support, posing a significant challenge for traffic forecasting. 
Recognizing this challenge, we have made a noteworthy observation: traffic patterns exhibit similarities across diverse cities.
Building on this key insight, we propose a solution for the cross-city few-shot traffic forecasting problem called \textbf{\underline{M}}ulti-scale \textbf{\underline{T}}raffic \textbf{\underline{P}}attern \textbf{\underline{B}}ank (\textbf{MTPB}).
Primarily, MTPB initiates its learning process by leveraging data-rich source cities, effectively acquiring comprehensive traffic knowledge through a spatial-temporal-aware pre-training process.
Subsequently, the framework employs advanced clustering techniques to systematically generate a multi-scale traffic pattern bank derived from the learned knowledge.
Next, the traffic data of the data-scarce target city could query the traffic pattern bank, facilitating the aggregation of meta-knowledge.
This meta-knowledge, in turn, assumes a pivotal role as a robust guide in subsequent processes involving graph reconstruction and forecasting.
Empirical assessments conducted on real-world traffic datasets affirm the superior performance of MTPB, surpassing existing methods across various categories and exhibiting numerous attributes conducive to the advancement of cross-city few-shot forecasting methodologies.
The code is available in~\url{https://github.com/zhyliu00/MTPB}.

\end{abstract}

}

\maketitle
\definecolor{lgray}{rgb}{0.9,0.9,0.9}

\section{Introduction}

\begin{figure*}
    \centering
    \includegraphics[width=0.85\linewidth]{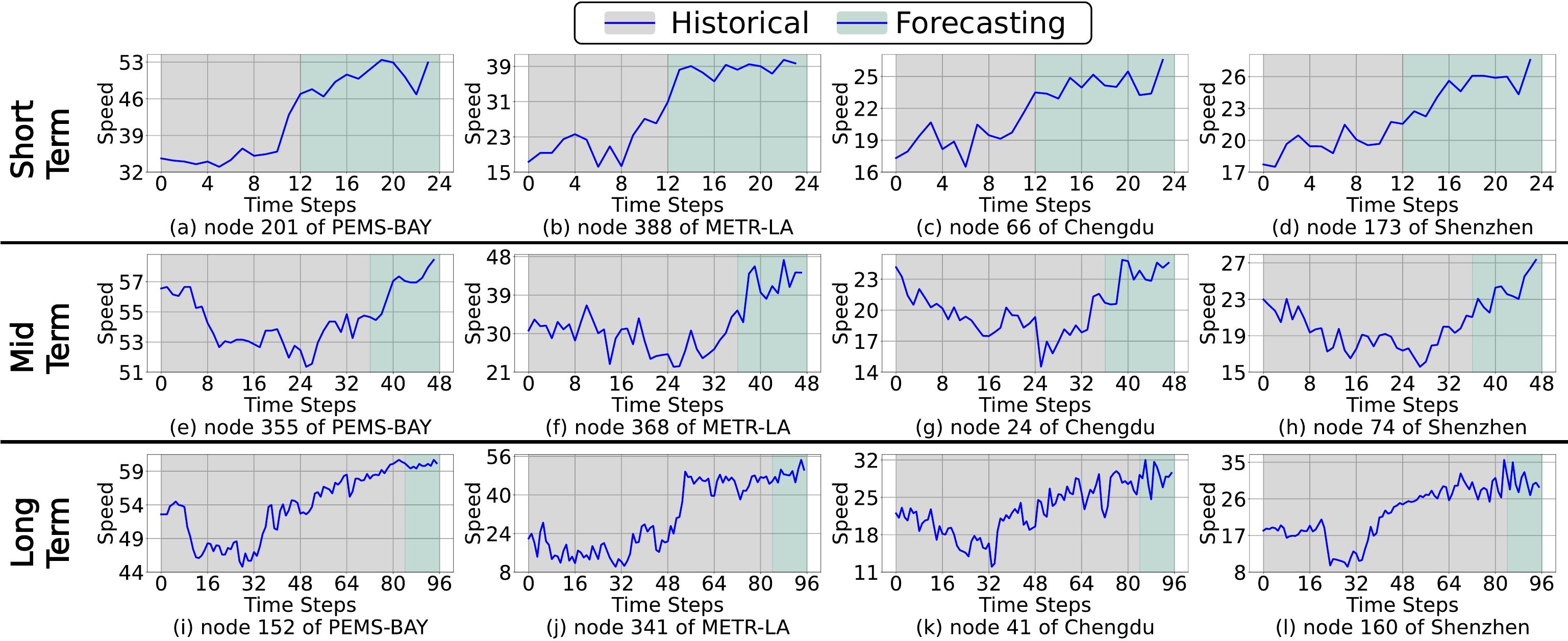}
    \caption{Multi-scale Traffic Patterns. (a)$\sim$(d) A rapid increase in speed is indicative of the future speed continuing to increase.
    (e)$\sim$(h) A nearby drop can lead to a subsequent rapid increase in the future.
    (i)$\sim$(l) If the speed has recently undergone a significant increase, the future speed is likely to exhibit fluctuations.
    }
    \label{fig:1intro_fig}
    \vspace{-.3cm}
\end{figure*}

\IEEEPARstart{T}{raffic} forecasting has emerged as a pivotal service within the realm of Intelligent Transportation Systems (ITS).
It aims to utilize the historical traffic data to forecast future traffic.
By accurately predicting the future traffic, various various downstream applications could be enabled such as route planning~\cite{li2019traffic}, traffic signal control~\cite{wei2018intellilight}, and traffic tolling~\cite{wang2022ctrl}.
These applications heavily rely on traffic data obtained from traffic sensors~\cite{li2017diffusion} or vehicle devices~\cite{Didi}.

In developed cities, these devices are well-deployed, and data collection is relatively straightforward. 
However, the situation differs in developing cities, where the availability of devices may be limited, leading to a scarcity of collected data. 
Consequently, training a deep model capable of capturing comprehensive traffic knowledge and accurately forecasting future traffic based on the limited data becomes a challenging task.
As a potential solution to this issue, it holds promise to learn robust traffic knowledge from data-rich cities and transfer that knowledge to data-scarce cities, thereby addressing the cross-city few-shot traffic forecasting problem.

The crux of the cross-city few-shot traffic forecasting problem lies in the definition and acquisition of robust traffic knowledge that possesses transferability across various cities.
In recent years, a considerable body of research has been dedicated to addressing this problem.
RegionTrans~\cite{wang2018cross} and CrossTRes~\cite{jin2022selective} perceive the implicit regional correlation between data-rich cities and data-scarce city, particularly in terms of their grid-based representation, as the foundation for robust traffic knowledge.
MetaST~\cite{yao2019learning} employs a global learnable traffic memory mechanism to learn robust traffic knowledge from data-rich cities.
However, these approaches rely on auxiliary data, such as event information, for knowledge transfer, which proves challenging when such data is unavailable in the data-scarce city.
ST-MetaNet~\cite{pan2019urban} and ST-GFSL~\cite{lu2022spatio} adopt the learned meta-knowledge to generate the parameters of spatial-temporal neural networks and subsequently employ these networks to forecast future traffic data in data-scarce cities.
However, these methods implicitly learn the meta-knowledge, overlooking the pronounced and explicit correlation that exists between the traffic data of different cities, resulting in reduced effectiveness.
TransGTR~\cite{jin2023transferable} aims to learn the city-agnostic node features and transferable graph structure between cities and then apply the model to the target city.
Nonetheless, TransGTR is limited in its ability that it can only handle one source city and one target city at a time.
Consequently, it falls short of effectively capturing and leveraging the robust shared traffic knowledge from multiple cities, thereby diminishing its overall effectiveness.

In fact, different cities have similar traffic patterns.
Namely, when the historical traffic status of roads of multiple cities exhibits similarities, it is indicative that their future traffic is also likely to be similar.
Moreover, this phenomenon could be observed not only in the short term but also in the long term as shown in Fig.~\ref{fig:1intro_fig}, where we plot the traffic speed of four cities as an example.
Fig.~\ref{fig:1intro_fig}(a)$\sim$(d) shows the 1-hour short-term traffic pattern that a rapid increase of speed manifests the future speed will continue to increase. 
Fig.~\ref{fig:1intro_fig}(e)$\sim$(h) shows the 3-hour mid-term traffic pattern that a nearby drop could lead to a rapid increase in the future. 
Fig.~\ref{fig:1intro_fig}(i)$\sim$(l) shows the 7-hour long-term traffic pattern that the future speed would fluctuate for a while if the speed has just rapidly increased.
Considering these multi-scale traffic patterns, it becomes evident that capturing and leveraging such robust traffic knowledge can significantly enhance the performance of cross-city few-shot traffic forecasting models.

Inspired by this observation, we proposed a \textbf{M}ulti-scale \textbf{T}raffic \textbf{P}attern \textbf{B}ank framework for cross-city few-shot traffic forecasting, which is abbreviated as \textbf{MTPB}. 
Specifically, it contains four modules: the Pre-training module, the Multi-scale Pattern Generation module, the Pattern Aggregation module, and the Forecasting module.
In the Pre-training module, the full data of the source data-rich cities is utilized to pre-train a robust traffic encoder.
The encoder aggregates the spatial-temporal information and aims to reconstruct the masked input data based on the idea of MAE~\cite{he2022masked}.
In the Multi-scale Pattern Generation module, the pre-trained encoder projects the data of the source data-rich cities to high-dimensional space, and the embeddings are clustered to construct the multi-sale traffic pattern bank.
In the Pattern Aggregation module, the data of the target data-scarce city could query the multi-scale traffic pattern bank and get the meta-knowledge.
Then, the graph structure is reconstructed based on the self-expressiveness of the meta-knowledge~\cite{kang2022fine,xu2019scaled} and used in the short-term spatial-temporal model.
Finally, in the Forecasting module, the meta-knowledge is concatenated with the output of short-term and long-term models and used to forecast future traffic.
The \textit{Reptile} meta-learning framework~\cite{nichol2018first} is utilized to get a better initialization of the model parameters of Pattern Aggregation and Forecasting modules.
Overall, the MTPB framework achieves 17.17\%, 20.54\%, and 20.95\% performance improvement in RMSE, MAE, and MAPE over state-of-the-art baselines on average respectively.

We summarize our contributions as follows.
\begin{itemize}
    \item We systematically investigate the cross-city few-shot traffic forecasting problem, where we highlight the significance of similar multi-scale traffic patterns across cities in accurately forecasting the future traffic in the data-scarce city.
    \item We propose a novel cross-city few-shot traffic forecasting framework based on multi-scale traffic pattern bank MTPB. It effectively harnesses the robust and explicit correlations of traffic patterns between different cities, leading to improved predictions of future traffic in the data-scarce city.
    \item We demonstrate the effectiveness of the MTPB framework through extensive experiments on real-world traffic datasets. The experimental results show that MTPB not only has achieved state-of-the-art performance but also shows characteristics that benefit the cross-city knowledge transfer problem.
\end{itemize}

\section{Related Work}
\noindent
\textbf{Traffic Forecasting.}
Traffic forecasting has gained considerable attention due to its practical applications.
Before deep learning, traditional approaches, such as those employing SVM\cite{nikravesh2016mobile}, probabilistic models~\cite{akagi2018fast}, or simulation methods~\cite{liang2022cblab}, have demonstrated satisfactory outcomes.
Recently, the prosperity of GCN, RNN, and Attention contributes to the deep learning methods for modeling the spatial-temporal traffic graph thus benefiting the area of traffic forecasting.
DCRNN~\cite{li2017diffusion}, STGCN~\cite{yu2017spatio}, STDN~\cite{yao2019revisiting}, STFGNN~\cite{li2021spatial},
Frigate~\cite{gupta2023frigate},
HIEST~\cite{ma2023rethinking} combine modules such as GCN, GRU, and LSTM to model the spatial-temporal relation. 
To better capture the dynamic spatial-temporal relations of the nodes of the traffic graph, methodologies such as AGCRN~\cite{bai2020adaptive}, Graph Wavenet~\cite{wu2019graph}, GMAN~\cite{zheng2020gman}, D2STGNN~\cite{shao2022decoupled}, ST-WA~\cite{cirstea2022towards}, DSTAGNN~\cite{lan2022dstagnn},
TrendGCN~\cite{jiang2023enhancing},
MC-STL~\cite{zhang2023mask} utilize techniques such as attention to reconstruct the adaptive adjacent matrix and fuse the temporal long-term relation to make better predictions.
\cite{duan2023localised} propose to use a localized adaptive graph to improve the performance.
MTGNN~\cite{wu2020connecting} and DMSTGCN~\cite{han2021dynamic} utilize auxiliary information that helps forecast the traffic.
STG-NCDE~\cite{choi2022graph}, STDEN~\cite{ji2022stden} model the traffic based on ordinary differential equation(ODE).
DGCNN~\cite{diao2019dynamic}, StemGNN~\cite{cao2020spectral} view the traffic forecasting task in a graph spectral view.
PM-MemNet~\cite{lee2021learning} learns and clusters the traffic flow patterns.
STEP~\cite{shao2022pre} adapts MAE~\cite{he2022masked} and proposes a pipeline to pre-train a model.
FDTI~\cite{liu2023fdti} builds layer graphs to conduct fine-grained traffic forecasting.
PECPM~\cite{wang2023pattern} tries to find the traffic pattern from the historical raw data.
STGBN~\cite{fan2023spatial} utilizes gradient boosting to enhance the model.
The existing methods discussed above primarily concentrate on single-city traffic forecasting. However, when it comes to cross-city few-shot traffic forecasting, these approaches face challenges such as distribution shift and over-fitting to one city. Moreover, certain modules, such as node embedding, fail to effectively handle the complexity of multiple cities and result in subpar performance.

\noindent
\textbf{Cross-City Few-Shot Learning.}
Few-Shot Learning has demonstrated promising results in various domains such as computer vision~\cite{snell2017prototypical}, natural language processing~\cite{lee2022meta}, and reinforcement learning~\cite{finn2017model} when facing the problem of data scarcity and distribution shift.
In the area of urban computing, some methods aim to solve city data scarcity by transferring city knowledge.
Floral~\cite{wei2016transfer} transfers the knowledge of rich multimodal data and labels to the target city to conduct air quality prediction, which is mainly designed for classification problems.
RegionTrans~\cite{wang2018cross} and CrossTReS~\cite{jin2022selective} learn the region correlation between the source cities to the target city.
MetaST~\cite{yao2019learning} learns a global memory which is then queried by the target region.
STrans-GAN~\cite{zhang2022strans} utilizes the GAN-based model to generate future traffic based on traffic demand.
However, these approaches mainly focus on grid-based data pertaining to city regions with multimodal auxiliary information, rather than graph-based univariate urban data.
ST-MetaNet~\cite{pan2019urban} and ST-GFSL~\cite{lu2022spatio} generate the parameters of spatial-temporal neural networks according to the learned meta-knowledge.
However, these methods essentially propose alternative parameter initialization strategies, without fully exploiting the rich information and strong correlations inherent in traffic patterns across different cities.
TransGTR~\cite{jin2023transferable} addresses the challenge of cross-city few-shot traffic forecasting by learning city-agnostic features and transferable graphs from one city to another city.
However, its limitation lies in its inability to capture diverse and robust traffic information originating from multiple cities.

\section{Preliminary}
\begin{table}[!t]
    \centering
    \caption{Table of important notations.}
    \vspace{-.3cm}
    \resizebox{0.85\linewidth}{!}{
    \begin{tabular}{c|c}
    \toprule
        Variable & Definition \\
        \midrule
        \midrule
        $T_0$ & The length of the input historical data\\
        $T'$ & The length of the forecasting data\\
        $P$ & The length of each traffic patch\\
        $s$ & The number of scale of traffic patterns\\
        $\mathbf{X}^i_j$ & The traffic data at time $j$ of node $i$\\
        $\mathbf{S}^i_j$ & The traffic patch at time $j$ of node $i$\\
        $\mathbf{B}$ & The multi-scale traffic pattern bank\\
        $\mathcal{Y}, \hat{\mathcal{Y}}$ & The ground truth and predicted future data\\
        $\mathbf{Z}$ & The meta-knowledge\\
        $\mathbf{A}$ & The adjacent matrix of raw data\\
        $\mathbf{A'}$ & The adjacent matrix of inner product result of $\mathbf{Z}$\\
        $\mathbf{C}$ & The coefficient matrix\\
        $\mathbf{\hat{A}}$ & The reconstructed adjacent matrix\\
        $\mathbf{R}_s,\mathbf{R}_l$ & The short-term and long-term embedding\\
        $\mathbf{H}^i_j$ & The hidden states of traffic patch $\mathbf{S}^i_j$\\
    \bottomrule
    \end{tabular}
    }
    \vspace{-.4cm}
    \label{tab:Definition}
\end{table}

\begin{definition}
\textbf{Traffic Spatio-Temporal Graph:} A traffic spatio-temporal graph can be denoted as $\mathcal{G}=(\mathcal{V},\mathcal{E},\mathbf{A},\mathbf{X})$. $\mathcal{V}$ is the set of nodes and $N=|\mathcal{V}|$ is the number of nodes. $\mathcal{E}$ is the set of edges and each edge can be denoted as $e_{ij}=(v_i,v_j)$. $\mathbf{A} \in \mathbb R^{N\times N}$ is the adjacency matrix of $\mathcal{G}$, where ${a_{ij}}=1$ indicates there is an edge between $v_i$ and $v_j$. By denoting $T_{total}$ as the total number of time steps, $\mathbf{X}\in\mathbb{R}^{N\times T_{total}\times C}$ represents the node feature that contains the traffic time series data, such as traffic speed, traffic volume, and time of day.
Here, $C$ indicates the input channel and $\mathbf{X}_t\in\mathbb{R}^{N\times C}$ represents the traffic data at time step $t$.
In this paper, we mainly focus on the public traffic speed dataset.
\end{definition}
\begin{definition}
\textbf{Traffic Patch \& Pattern:} \textit{Traffic patch} refers to a fixed interval of traffic time series data.
To illustrate, if we have traffic data for a full day, we can divide it into 24 one-hour traffic patches.
Formally, given the traffic time series data denoted as $\mathbf{X}^i$ for a specific location $v_i$, a traffic patch is represented as $\mathbf{S}^i_t := \mathbf{X}_{t:t+P}^i\in \mathbb{R}^{P\times C}$, where $P$ denotes the length of a patch and $C$ represents the number of features captured in each time step.
In this paper, the temporal granularity of the data is five minutes, i.e., the data is sampled every five minutes.
We set each patch to contain 12 time steps, i.e., $P=12$.
By doing so, each traffic patch represents the traffic data for one hour.

\textit{Traffic pattern} refers to a representative traffic patch encapsulating robust and transferable information across different cities.
When provided with a vast amount of raw traffic data in the form of traffic patches, the traffic patterns can be derived through a refinement process.
\end{definition}
\begin{problem}
\textbf{Traffic Forecasting:} Given the historical traffic data of $T_0$ steps, the goal of the traffic forecasting problem is to learn a function $f(\cdot)$ that forecasts the future traffic data of $T'$ steps. This task is formulated as follows.
\begin{equation}
    [\mathbf{X}_{t-T_0+1},\cdots,\mathbf{X}_{t}]\stackrel{f(\cdot)}{\longrightarrow}[\mathbf{X}_{t+1},\cdots,\mathbf{X}_{t+T'}]
\end{equation}
\end{problem}
\begin{problem}    
\textbf{Cross-city Few-Shot Traffic Forecasting:} Given $P$ source cities $\mathcal{G}^{source}=\{\mathcal{G}^{source}_1,\cdots,\mathcal{G}^{source}_{P}\}$ with a large amount of traffic data and a target city $\mathcal{G}^{target}$ with only a few traffic data, the goal of cross-city few-shot traffic forecasting is to learn a model based on the available data of both $\mathcal{G}^{source}$ and $\mathcal{G}^{target}$ to conduct traffic forecasting on the future data of $\mathcal{G}^{target}$.
  
\end{problem}

\section{Methodology}
\begin{figure*}[!th]
    \centering
    \includegraphics[width=0.95\linewidth]{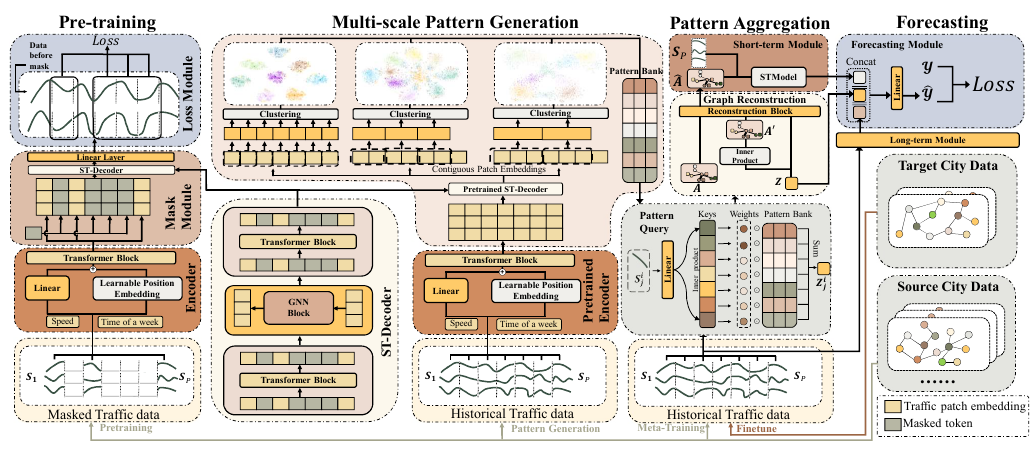}
    \centering
    \caption{Diagrams of MTPB. 1) In the Pre-training stage, a traffic patch encoder is pre-trained by the data of source cities. 2) In the Multi-scale Pattern Generation stage, source city traffic patches are processed by the pre-trained encoder, and the output embeddings are clustered to form the traffic pattern bank. 3) In the Pattern Aggregation stage, the traffic pattern bank is queried by the input traffic patches, and an adjacency matrix is reconstructed. 4) In the Forecasting stage, the metaknowledge along with the short-term and long-term models forecasts the future traffic.}
    \label{fig:main_fig}
    \vspace{-.3cm}
\end{figure*}

In this section, we will present our framework called MTPB (Multi-scale Traffic Pattern Bank) for the cross-city few-shot traffic forecasting task. The framework consists of four modules, as illustrated in Fig.~\ref{fig:main_fig}.
The first module is the Spatial-Temporal-Aware Pre-training module.
Unlike previous methods~\cite{liu2023cross,shao2022pre}, we incorporate spatial information in the pre-training process using a Spatial-Temporal Decoder (ST-Decoder) to pre-train a robust traffic data encoder.
Next, in the Multi-scale Pattern Generation module, the raw traffic data is fed into the pre-trained traffic patch encoder to obtain traffic patch embeddings and clustering is performed on the segmented embeddings to construct a multi-scale traffic pattern bank.
Subsequently, in the Pattern Aggregation module, the raw data from the target data-scarce city can directly query the multi-scale traffic pattern bank to obtain the condensed meta-knowledge. 
This meta-knowledge is then utilized to reconstruct the graph and assist in forecasting future traffic.
Finally, in the Forecasting module, the meta-knowledge, along with the outputs of a short-term forecasting model and a long-term forecasting model, are fused to generate the final forecasting result. 

\subsection{Spatial-Temporal-Aware Pre-training}

{\textbf{\textit{Goal}:}} 
Raw traffic patch data from source cities holds valuable information but is prone to noise introduced during data collection, aggregation, and splitting.
Moreover, existing in a low-dimensional space limits its expressiveness for downstream tasks.
To address this, we propose a self-supervised traffic patch encoder pre-training framework. 
This framework denoises and projects the data into a high-dimensional space, enhancing its expressiveness.
In this regard, we propose a spatial-temporal-aware pre-training framework for the traffic patch encoder. Unlike previous methods~\cite{liu2023cross,shao2022pre}, our framework incorporates both spatial and temporal information to generate traffic patch embeddings. 
This is in contrast to approaches that solely rely on temporal information. 

{\textbf{\textit{Masked Input:}}}
The input of the pre-train module is the raw city data of a given length $X_{t:t+T_0}$ and its corresponding adjacency matrix $A$.
The raw data is then segmented into many traffic patches $\mathbf{X}^i_{t:t+T_0}=\{\mathbf{S}^i_t, \mathbf{S}^i_{t+1},\cdots,\mathbf{S}^i_{t+{T_0/P}}\}$ where each traffic patch has length $P$ that $\mathbf{S}^i_t := \mathbf{X}_{t:t+P}^i$.
In this paper, $T_0$ is set to 288 to represent one-day data, and $P$ is set to 12 to represent one-hour data.
Next, we randomly select a subset of traffic patches and apply a masking operation to them.
The selected patches do not need to be contiguous or sequential.
Following previous approches~\cite{shao2022pre,he2022masked,liu2023cross}, we set the mask ratio to 75\%, which means 75\% of the traffic patch information is masked.
Such a high mask ratio makes the reconstruction task difficult so the pre-trained encoder is expected to be more robust and potentially more effective when transferred to other cities.

{\textbf{\textit{Encoder:}}}
The aim of the encoder is to aggregate the information of the unmasked traffic patches.
However, since the majority of the traffic patches are masked and the remaining unmasked patches are not contiguous, traditional CNN-based and RNN-based encoders are not well-suited for this task.
As a result, we opt to use a Transformer block~\cite{vaswani2017attention} as the encoder in our approach.
Prior to being input to the Transformer encoder, the unmasked traffic patch undergoes preprocessing steps.
Firstly, it is fed into a linear layer and combined with a learnable positional embedding. 
The positional embedding is based on the time of the week, and since there are 24 hours in a day and 7 days in a week, a total of 168 learnable positional embeddings are employed.
Once the positional embedding is added, the unmasked patches are concatenated and fed into the Transformer encoder.
The encoder generates corresponding output embeddings, capturing the relevant information from the traffic patches.
Formally, we can write as follows
\begin{equation}
    \mathbf{H}^i_j =  TS_E(\text{Concat}\{(\mathbf{W}_{enc}\mathbf{S}^i_j+\mathbf{b}_{enc})+\text{PE}(\mathbf{S}^i_j)|\mathsf{M}^i_j = 0\})
\end{equation}.
Here, $\text{PE}(\cdot)$ indicates learnable positional embedding, and $TS_E(\cdot)$ indicates the encoder transformer layers.
$\mathsf{M}^i_j=0$ indicates $\mathbf{S}^i_j$ is not masked and vice versa. 
$\mathbf{W}_{enc}\in\mathbb{R}^{d \times {PC}}$ and $\mathbf{b}_{enc}\in\mathbb{R}^d$ are learnable parameters.

{\textbf{\textit{Mask Module:}}}
The primary objective of this module is to aggregate spatial and temporal information from both the masked and unmasked traffic patches. 
The process involves several steps.
Firstly, a learnable token $\mathbf{H}_{mt}\in\mathbb{R}^{d}$ is created to represent the masked traffic patch. This token is duplicated and used to fill the positions of the masked patches.
Next, the duplicated token and the embeddings of the unmasked patches are concatenated. The resulting concatenated representation is then fed into a Spatial-Temporal Decoder (ST-Decoder).
The ST-Decoder consists of three steps. In the first and third steps, a Transformer block is employed to aggregate the temporal information. This helps capture the temporal patterns and dependencies within the traffic data.
In the second step, a Graph Neural Network (GNN) block is utilized to aggregate the spatial information. This allows the model to capture the spatial relationships and dependencies between different locations in the traffic data.
The output of the ST-Decoder, which incorporates both spatial and temporal information, is then passed through a linear layer. This linear layer generates the reconstructed masked traffic patch by mapping the aggregated information back to the original traffic patch space.
Formally, this process could be represented as follows
\begin{equation}
    \hat{\mathbf{S}}^i_j = \mathbf{W}_{dec}\cdot {ST_D(Concat\{\mathbf{H}^i_j,\mathbf{H}_{mt}\})} +\mathbf{b}_{dec}.
\end{equation}
Here, $ST_D(\cdot)$ indicates the ST-Decoder, and $\mathbf{W}_{dec}\in\mathbb{R}^{PC \times d }$ and $\mathbf{b}_{dec}\in\mathbb{R}^{PC}$ are learnable parameters.

{\textbf{\textit{Loss Module:}}}
In the pre-training stage, the primary task is to reconstruct the masked traffic patches, which constitute the majority of the data. 
Meanwhile, the unmasked patches serve as the input to the model. 
Consequently, the final loss function incorporates the Mean Square Error (MSE) of the reconstructed masked traffic patches.
\begin{equation}
    \mathcal{L}_{p}=\sum_{i=1}^N\sum_{j=1}^P \mathsf{M}^i_j(\mathbf{S}^i_{j} - \hat{\mathbf{S}}^i_{j})^2
\end{equation}

In summary, the pre-training stage focuses on reconstructing the masked traffic patches.
Through this process, the encoder and the ST-Decoder learn valuable features from the traffic patches and project them into a higher-dimensional denoised embedding space, leading to improved performance in subsequent tasks such as pattern generation and forecasting.

\subsection{Multi-scale Pattern Generation}

\begin{figure}[!t]
\centering
\resizebox{0.9\linewidth}{!}{
\removelatexerror
\label{alg:tpg}

\begin{algorithm}[H]
    \KwIn{Pre-trained traffic patch encoder $TS_E$ and $ST_D$, Historical traffic patch of source cities $\mathbf{S}^i_j$, Pattern scales $\{c_1, \cdots, c_s\}$, number of clusters $K$.}
    \KwOut{Traffic pattern bank $\mathbf{B}\in\mathbb{R}^{s\times K\times d}$}

    $\mathbf{H}_t,\cdots,\mathbf{H}_{t+T_0/P} = ST_D(TS_E(\{\mathbf{S}_t,\cdots,\mathbf{S}_{t+T_0/P}\}))$\;
    
    \For{$c \in \{c_1, \cdots, c_s\}$}{
        $\mathbf{H}^i_{t}, \cdots \xrightarrow{segment} \mathbf{H}^i_{t:t+c}, \cdots$

        Collect $\{\mathbf{H}^i_{t:t+c}|i=1,...,N\}$\;  
        
        $o^c_{1:K} = KMeans(\{\mathbf{H}^i_{t:t+c}|i=1,...,N\})$\;
        
        $\mathbf{B}^c\leftarrow o^c_{1:K}$\;
        
    }
    
    \Return $\mathbf{B}$\;

    \caption{Multi-scale Traffic Pattern Generation}
\end{algorithm}
}
\end{figure}

{\textbf{\textit{Goal}:}} 
The extensive traffic datasets within the source cities contain valuable information for few-shot traffic prediction in the target city.
However, directly employing the extensive traffic data for forecasting becomes impractical due to its sheer volume. 
As illustrated in Fig.~\ref{fig:1intro_fig}, the traffic data exhibits multi-scale patterns, inspiring the direction to reduce the data size. 
Consequently, we could encode the traffic patch by the pre-trained encoders and leverage clustering methodologies to pinpoint representative traffic patches, forming a condensed traffic pattern bank.
This refined traffic pattern bank not only alleviates the computational complexity of the framework but also serves as a concise robust representation of traffic patterns of the source cities.

{\textbf{\textit{Framework:}}}
We directly utilize the traffic patch encoder $TS_E(\cdot)$ and $ST_D(\cdot)$ that is pre-trained in the previous stage. 
The enormous traffic patches of all nodes in the source cities $\mathbf{S}_j$ are put into the patch encoder to generate traffic patch embeddings $\mathbf{H}_j$. 
\begin{equation}
    \label{eq:Enc}
    \mathbf{H}_t,\cdots,\mathbf{H}_{t+T_0/P} = ST_D(TS_E(\{\mathbf{S}_t,\cdots,\mathbf{S}_{t+T_0/P}\}))
\end{equation}
For simplicity, the linear layers are omitted here.
The pre-trained encoders could project the traffic patches into high-dimensional denoised embedding space, and traffic patches with similar traffic semantics are projected to near space.

Next, we aim to generate multi-scale traffic patterns.
The scale is the number of contiguous traffic patch that comprises a traffic pattern.
We first set the scales of the to-be-clustered traffic patterns as $\{1, 3, 6, 12, 24\}$, which means one hour, three hours, and so on.
Then, for each scale $c$, the traffic patch embeddings for each node in the source cities $\mathbf{H}^i_j$ are segmented based on the scale.
\begin{equation}
    \mathbf{H}^i_{t}, \cdots, \mathbf{H}^i_{t+T_0/P} \xrightarrow{segment} \mathbf{H}^i_{t:t+c}, \cdots, \mathbf{H}^i_{t+T_0/P-c:t+T_0/P}
\end{equation}

Subsequently, we can then use clustering techniques to group similar segmented traffic patch embeddings.
Various methods can be employed for clustering segmented traffic patch embeddings, encompassing unsupervised techniques such as k-means and density-based clustering. 
In this context, to streamline the framework and obtain resilient traffic patterns, we opt for the use of the \textit{k-means} clustering algorithm, employing the \textit{cosine} distance function.

\begin{equation}
    \label{eq:KMeans}
    o^c_{1:K} = KMeans(\{\mathbf{H}^i_{t:t+c}|i=1,...,N\})
\end{equation}
Here, $o^c_{1:K}\in\mathbb{R}^{K\times d}$ are the centroids of the clusters.

After grouping the traffic patch embeddings into clusters, a representative set of traffic patterns is chosen from each cluster to build the traffic pattern bank. 
In the context of few-shot traffic forecasting, where the traffic patches in the target city are unknown and potential distribution bias may be present, selecting centroids becomes crucial. 
The centroid, in this context, signifies the overall characteristics of the traffic patterns within the cluster rather than being specific to any individual traffic patch.
Therefore, we designate the centroids of each cluster $o^c_{1:K}$ as the traffic pattern bank of scale $c$ to ensure robustness and transferability, which is noted as $\mathbf{B}^c\leftarrow o^c_{1:K}\in\mathbb{R}^{K\times d}$.

\begin{figure}[!t]
\centering
\resizebox{0.9\linewidth}{!}{
\removelatexerror
\label{alg:meta}

\begin{algorithm}[H]
    \KwIn{Forecasting model $F_\theta(\cdot)$, Source city data $G_{source}$, Target city data $G_{target}$}
    \KwOut{Trained Model Parameter $\theta$}
    \tcc{Meta-training the $\theta$ with $G_{source}$}
    Random initialize $\theta$\;
    
    \For{$e\leftarrow range(0,meta\_epochs,1)$}{
        $\tau_{spt}, \tau_{qry}\longleftarrow SampleTask(G_{source})$\;
        
        $\mathbf{S}_{spt}, \mathcal{Y}_{spt}\longleftarrow \tau_{spt}$\;  
        
        $\mathbf{S}_{qry}, \mathcal{Y}_{qry}\longleftarrow \tau_{qry}$\;
        $\theta_{\tau}\longleftarrow\theta$\;
        
        \For{$i\leftarrow range(0,update\_step,1)$}{
            $\hat{\mathcal{Y}}^{\theta_\tau}_{spt}\longleftarrow F_{\theta_\tau}(S_{spt})$\;
            
            compute $\triangledown_{\theta_\tau}\mathcal{L}_i(\hat{\mathcal{Y}}^{\theta_\tau}_{spt}, \mathcal{Y}_{spt})$ by Eq~\eqref{eq:loss}\;
            
            ${\theta_\tau}\longleftarrow {\theta_\tau} - \alpha\triangledown_{\theta_\tau}\mathcal{L}_i(\hat{\mathcal{Y}}^{\theta_\tau}_{spt}, \mathcal{Y}_{spt})$\;
            $\hat{\mathcal{Y}}^{\theta_\tau}_{qry}\longleftarrow F_{\theta_\tau}(S_{qry})$\;
            
            compute \& store $\triangledown_{\theta_\tau}\mathcal{L}_i(\hat{\mathcal{Y}}^{\theta_\tau}_{qry}, \mathcal{Y}_{qry})$\;
            
        }
        \tcp{use the gradients to update $\theta$}
        \For{$i\leftarrow range(0,update\_step,1)$}{
            $\theta\longleftarrow\theta - \frac{\beta}{update\_step}\triangledown_{\theta_\tau}\mathcal{L}_i(\hat{\mathcal{Y}}^{\theta_\tau}_{qry}, \mathcal{Y}_{qry})$
        }
    }  
    \tcc{Finetune the $\theta$ with $G_{target}$}
    \For{$e\leftarrow range(0,train\_epochs,1)$}{
        \For{Iterate All Batches}{
            $\mathbf{S}, \mathcal{Y}\longleftarrow SampleBatch(G_{target})$\;
            $\hat{\mathcal{Y}}^{\theta}\longleftarrow F_{\theta}(\mathbf{S})$\;
            compute $\triangledown_{\theta}\mathcal{L}(\hat{\mathcal{Y}}^{\theta}, \mathcal{Y})$ by Eq~\eqref{eq:loss}\;
            use $\triangledown_{\theta}\mathcal{L}(\hat{\mathcal{Y}}^{\theta}, \mathcal{Y})$ to update $\theta$\;
        }
    }
    \Return $\theta$

\caption{Meta-training and Fine-tuning}
\end{algorithm}
    }
    \vspace{-.4cm}
\end{figure}

\subsection{Pattern Aggregation}

{\textbf{\textit{Goal}:}}
The goal of Pattern Aggregation is to aggregate the meta-knowledge from the multi-scale traffic pattern bank.
The core idea is to query the multi-scale traffic pattern bank with historical traffic patch embeddings and aggregate the retrieved patterns as the auxiliary metaknowledge.
The metaknowledge is used to reconstruct the graph structure and guide the downstream short-term model to forecast future traffic.

{\textbf{\textit{Pattern Query}:}}
The traffic pattern bank comprises two components: \textit{Key} and Pattern Bank $\mathbf{B}$.
The \textit{Key} is a learnable embedding matrix with dimensions $\mathbb{R}^{s\times K\times d_q}$, where $s$ denotes the number of scales, $K$ is the number of patterns, and $d_q$ represents the embedding shape of \textit{Key}. 
The Pattern Bank $\mathbf{B}\in\mathbb{R}^{s\times K\times d}$ is a fixed embedding matrix from the preceding clustering stage.

For a given query, denoted as $\mathbf{S}^i={\mathbf{S}_1^i,\cdots,\mathbf{S}_{T_0/P}^i}$, consisting of $P$ continuous traffic patches of $v_i$, the raw traffic patches are initially projected into the embedding space of \textit{Key}. 
Subsequently, dot product scores $\omega$ are computed between each linearly transformed query patch and each key in the \textit{Key} matrix, expressed as follows
\begin{equation}
    \omega^c_k = dot(\mathbf{W}^c\mathbf{S}^i_j+\mathbf{b}^c,Key^c_k).
\end{equation}
Here, $\mathbf{W}^c\in\mathbb{R}^{d_q\times d}$ and $\mathbf{b}^c\in\mathbb{R}^{d_q}$ represents the linear layer of scale $c$. $Key^c_k\in\mathbb{R}^{d_q}$ is the \textit{k-th} key of the traffic pattern of scale $c$ and $\omega^c_k\in\mathbb{R}$ is the score. 
Then, the retrieved traffic pattern $\mathbf{Z}^{c,i}_j\in\mathbb{R}^d$ could be represented as a weighted sum of multi-scale traffic pattern $\mathbf{B}^c_k\in\mathbb{R}^d$ as follows
\begin{equation}
    \mathbf{Z}^{c,i}_j=\sum_{k=1}^K \omega^c_k\cdot \mathbf{B}^c_k.
\end{equation}
Finally, we use a Transformer layer to aggregate the meta-knowledge embedding sequence of each scale and fuse the multi-scale meta-knowledge with a linear layer.
\begin{equation}
    \mathbf{Z}^{i,c}=TS_s\{\mathbf{Z}^{i,c}_0,\cdots, \mathbf{Z}^{i,c}_{T_0/P}\})
\end{equation}
\begin{equation}
    \mathbf{Z}^i=Linear(Concat\{\mathbf{Z}^{i,0},\cdots, \mathbf{Z}^{i,c}\})
\end{equation}
Where $\mathbf{Z}^i\in\mathbb{R}^d$ is the meta-knowledge of node $i$.
This process enables the model to retrieve the most relevant traffic patterns for a given query patch, thereby generating robust meta-knowledge from source cities.

{\textbf{\textit{Self-expressive Graph Reconstruction}:}}
To fully exploit the structural information contained in the meta-knowledge and denoise the raw adjacency matrix, we propose to use the self-expressive method~\cite{kang2022fine,xu2019scaled} to fuse the multi-source structural information.
The self-reconstruction method aims to generate a coefficient matrix $\mathbf{C}\in\mathbb{R}^{n\times n}$ that reconstruct the meta-knowledge $\mathbf{Z}\in\mathbb{R}^{n\times d}$
\begin{equation}
    \mathbf{C} = \arg\min_{\mathbf{C}}||\mathbf{Z}^T-\mathbf{Z}^T\mathbf{C}||_F^2
\end{equation}
After the optimization, $\mathbf{C}$ could represent the similarity between the metaknowledge of each node and thus represent the denoised graph structure. 
However, the optimization could easily fall into a trivial solution that $\mathbf{C}=I$.
To avoid this, we need to add a regularization term $\Theta(\mathbf{C})$ as follows
\begin{equation}
    \mathbf{C} = \arg\min_{\mathbf{C}}||\mathbf{Z}^T-\mathbf{Z}^T\mathbf{C}||_F^2 + \gamma \Theta(\mathbf{C}).
\end{equation}
Considering that we now have the structural information contained in the raw adjacent matrix $\mathbf{A}$ and the inner-product of the linear-transformed meta-knowledge $\mathbf{A'}=Softmax(Linear_Q(\mathbf{Z})Linear_K(\mathbf{Z}^T))$, we could use them as the regularization term to fuse the information into the coefficient matrix $\mathbf{C}$.
\begin{equation}
    \mathbf{C} = \arg\min_{\mathbf{C}}||\mathbf{Z}^T-\mathbf{Z}^T\mathbf{C}||_F^2 + \gamma (||\mathbf{C}-\mathbf{A}|| + ||\mathbf{C}-\mathbf{A'}||)
\end{equation}
The equation could be easily solved by setting the first-order derivative w.r.t. $\mathbf{C}$ to zero as follows
\begin{equation}
    -\mathbf{Z}(\mathbf{Z}^T-\mathbf{Z}^T\mathbf{C})+\gamma(\mathbf{C}-\mathbf{A})+\gamma(\mathbf{C}-\mathbf{A'})=0.
\end{equation}
\begin{equation}
    \mathbf{C}=(\mathbf{Z}\mathbf{Z}^T+2\gamma I)^{-1}(\mathbf{Z}\mathbf{Z}^T+\gamma(\mathbf{A}+\mathbf{A'}))
\end{equation}
We add an additional step to make the adjacent matrix symmetry.
\begin{equation}
    \mathbf{\hat{A}}=\frac{\mathbf{C}+\mathbf{C}^T}{2}
\end{equation}

\subsection{Forecasting}

{\textbf{\textit{Goal}:}}
The goal of the Forecasting module is to employ the meta-knowledge $\mathbf{Z}$ and the reconstructed adjacency matrix $\mathbf{\hat{A}}$ to aid in forecasting future traffic.
Simultaneously, a meta-learning-based method is utilized to obtain improved initial parameters for the forecasting models.

{\textbf{\textit{Short-term Module}:}}
The short-term module consists of a spatial-temporal model (STModel). 
The STModel receives input from short-term traffic patches $\mathbf{S_t}\in\mathbb{R}^{N\times P\times C}$ and the reconstructed adjacency matrix $\mathbf{\hat{A}}$, extracting information to produce short-term embedding $\mathbf{R}_s=STModel(\mathbf{S_t},\mathbf{\hat{A}})\in\mathbb{R}^{N\times d}$.
Here we choose Graph Wavenet~\cite{wu2019graph} as the STModel.

{\textbf{\textit{Long-term Module}:}}
Due to linear becomes a strong baseline in the long-term forecasting task~\cite{zeng2023transformers,das2023long}, the long-term module here is composed of a linear layer that takes in historical long-term data $\mathbf{X}_{t-T_0:t}\in\mathbb{R}^{N\times T_0\times C}$.
The long-term embedding is then generated $\mathbf{R}_l=Linear(\mathbf{X}_{t-T_0:t})\in\mathbb{R}^{N\times d}$

{\textbf{\textit{Forecast}:}}
To further incorporate the metaknowledge into the forecasting process, we concatenate the metaknowledge $\mathbf{Z}\in\mathbb{R}^{N\times d}$ with the representation $\mathbf{R}_s,\mathbf{R}_l\in\mathbb{R}^{N\times d}$.
The resulting representation is then fed into a regression layer to generate the forecast traffic data $\hat{\mathcal{Y}}$, which is a Multi-Layer Perception (MLP).
This process can be represented as follows
\begin{equation}
    \begin{aligned}
        \hat{\mathcal{Y}} & =MLP([\mathbf{M}||\mathbf{R}_s||\mathbf{R}_l]).
    \end{aligned}
\end{equation}
Here, $\hat{\mathcal{Y}}\in\mathbb{R}^{N\times T'\times C}$ is the forecasting result. 
Given the ground truth $\mathcal{Y}\in\mathbb{R}^{N\times T'\times C}$, mean square error is selected as the loss.
\begin{equation}
    \label{eq:loss}
    \mathcal{L}=\frac{1}{NT'C}\sum_{i=1}^N\sum_{j=1}^{T'}\sum_{k=1}^{C}(\mathcal{Y}_{ijk}-\hat{\mathcal{Y}}_{ijk})^2
\end{equation}

{\textbf{\textit{Meta-Training \& Fine-tuning:}}}
Though the multi-scale traffic pattern bank contains rich metaknowledge about the traffic patches of source cities, other models need to adapt to the target city such as the short-term and long-term models.
To address this issue, we propose a meta-learning-based approach to learn better initial parameters from city data $G_{source}$.
We adopt \textit{Reptile}~\cite{nichol2018first} meta-learning framework here, which essentially conducts multi-step gradient descent on the query tasks.
Then, we fine-tune the model with the few-shot data of the target city $G_{target}$.
After that, the model parameters are adapted to the target city and the forecasting performs better.
Formally, by denoting the forecasting model as $F_\theta(\cdot)$ and its corresponding parameter as $\theta$, meta-training and fine-tuning are shown in Alg.~\ref{alg:meta}.

\section{Experiment}

This section presents a thorough evaluation of our proposed framework, MTPB. 
Specifically, We shall address the following research questions.
\begin{itemize}[leftmargin=*]
    \item \noindent{\textbf{RQ1:}} How does MTPB perform in comparison to other baselines in the task of few-shot traffic forecasting?
    \item \noindent{\textbf{RQ2:}} How does each component of MTPB contribute to the final forecasting performance?
    \item \noindent{\textbf{RQ3:}} Does the Pre-training module effectively capture robust knowledge from the source cities?
    \item \noindent{\textbf{RQ4:}} How can the clustering parameter \textit{K} be optimally chosen to obtain a proficient traffic pattern bank?
    \item \noindent{\textbf{RQ5:}} Is the Graph Reconstruction module successfully integrating structural information?
    \item \noindent{\textbf{RQ6:}} Can MTPB enhance the performance of various STmodels in the context of few-shot forecasting?
    \item \noindent{\textbf{RQ7:}} How does MTPB perform in efficiency?
\end{itemize}

\subsection{Experiment Settings}
\begin{table}[tp]
    \caption{Statistical details of traffic datasets.}
    \label{tab:data}
    \centering
    \resizebox{0.85\linewidth}{!}{
    \begin{tabular}{c|c|c|c|c}
    \toprule
         & PEMS-BAY & METR-LA & Chengdu & Shenzhen\\
         \midrule
         \# of Nodes & 325 & 207 & 524 & 627\\ 
         \# of Edges & 2,694 & 1,722 & 1,120 & 4,845\\
         Interval & 5 min & 5 min & 10 min & 10 min\\
         \# of Time Step & 52,116 & 34,272 & 17,280 & 17,280\\
         Mean & 61.7768 & 58.2749 & 29.0235 & 31.0092\\
         Std & 9.2852 & 13.1280 & 9.6620 & 10.9694\\ 
    \bottomrule
    \end{tabular}
    }
    \vspace{-.2cm}
\end{table}

\noindent{\textbf{Dataset:}}
We evaluate our proposed framework on four real-world public datasets: \textit{PEMS-BAY}, \textit{METR-LA}~\cite{li2017diffusion}, \textit{Chengdu}, \textit{Shenzhen}~\cite{Didi}.
These datasets contain months of traffic speed data and the details of these data are listed in Table~\ref{tab:data}.

\noindent{\textbf{Few-shot Setting:}}
We use a similar few-shot traffic forecasting setting to~\cite{lu2022spatio}.
The data of these four cities are divided into source data, few-shot target data, and test data.
Here, source data consists of data from three cities, while few-shot target data and test data consist of data from the target city.
For example, if \textit{PEMS-BAY} is selected as the target city, the full data of \textit{METR-LA}, \textit{Chengdu}, \textit{Shenzhen} constitutes source data.
Then, three-day data of \textit{PEMS-BAY} constitutes the few-shot target data, and the remaining data of \textit{PEMS-BAY} constitutes the test data.
We train MTPB on source data and few-shot target data, and evaluate it in the test data.

\noindent{\textbf{Details:}}
In Pre-training and Pattern Generation phases, we use $T_0=288$ and $P=12$, which means one-day data is divided into 24 patches to form a patch series.
In the Forecasting stage, we also use the same $T_0$ and $P$ to forecast the future 36 steps of data.
Besides, \textit{Chengdu} and \textit{Shenzhen} have a longer data interval so we conduct linear interpolate to align the positional embedding.
The mask ratio of Pre-training is set to 75\% according to~\cite{liu2023cross,shao2022pre}.
The learning rate of Pre-training is set to 0.0001 and the learning rate of Meta-training $\alpha$ and $\beta$ are both set to 0.0005.
The $\gamma$ is set to 10.
The Adam optimizer has a learning rate of 0.001 and a weight decay of 0.01.
The dimension of $H$ and $Key$ and the positional embedding size are set to 128.
The number of traffic patterns $K$ is set to 10.
The number of tasks of meta-training is set to 2.
The experiment is implemented by Pytorch 1.10.0 on RTX3090.
We evaluate the performances of Rooted Mean Square Error (RMSE), Mean Absolute Error (MAE), and Mean Absolute Percentage Error (MAPE).

\subsection{RQ1: Overall Performance}

\begin{table*}[htb]
\caption{Overall performance of few-shot traffic forecasting on \textit{PEMS-BAY}, \textit{METR-LA}, \textit{Chengdu}, and \textit{Shenzhen}. M,C,S$\rightarrow$PEMS-BAY means the source data is \textit{METR-LA}, \textit{Chengdu}, \textit{Shenzhen}, and the target data is \textit{PEMS-BAY}. The mean and standard deviation of the results in 5 runs is shown. In each column, the best result is highlighted in bold and grey, and the second-best result is underlined. Marker * and ** indicates the mean of the results is statistically significant  (* means t-test with p-value $<$ 0.05 and ** means t-test with p-value $<$ 0.01).}
\label{tab:performance}
\resizebox{1\linewidth}{!}{

\begin{tabular}{c|c|c c c|c c c|c c c|c c c}
\toprule

& & \multicolumn{3}{c}{10 min} & \multicolumn{3}{c}{60 min} & \multicolumn{3}{c}{120 min}& \multicolumn{3}{c}{180 min}\\
\cline{3-14}
 & &RMSE & MAE & MAPE (\%) & RMSE & MAE & MAPE (\%) & RMSE  & MAE & MAPE (\%) & RMSE & MAE & MAPE (\%) \\ 

\midrule
\midrule
\multirow{ 11}{*}{\rotatebox{90}{M,C,S$\rightarrow$PEMS-BAY}}& HA & 9.447 & 5.549 & 14.521 & 9.447 & 5.549 & 14.521& 9.447 & 5.549 & 14.521& 9.447 & 5.549 & 14.521  \\ 
 & DCRNN & 3.55$\pm$0.16 & 2.25$\pm$0.12 & 4.80$\pm$0.32 & 6.10$\pm$0.16 & 3.41$\pm$0.15 & 8.05$\pm$0.41 & 8.18$\pm$0.32 & 4.60$\pm$0.26 & 10.98$\pm$0.45 & 8.98$\pm$0.44 & 5.27$\pm$0.38 & 12.69$\pm$0.54
\\ 
 & GWN & 3.03$\pm$0.16 & 1.98$\pm$0.10 & 3.68$\pm$0.27 & 6.12$\pm$0.09 & \underline{3.19$\pm$0.06} & 8.02$\pm$0.27 & 8.09$\pm$0.14 & 4.48$\pm$0.08 & 10.99$\pm$0.27 & 8.70$\pm$0.23 & 4.95$\pm$0.18 & 12.85$\pm$0.25
 \\ 
 & STFGNN & 3.17$\pm$0.26 & 2.05$\pm$0.16 & 3.78$\pm$0.63 & 6.19$\pm$0.15 & 3.24$\pm$0.10 & 8.31$\pm$0.56 & 8.02$\pm$0.15 & 4.50$\pm$0.10 & 11.24$\pm$0.38 & 8.44$\pm$0.13 & 4.85$\pm$0.19 & 12.88$\pm$0.33
 \\ 
 & DSTAGNN & 3.30$\pm$0.42 & 2.22$\pm$0.57 & 4.47$\pm$0.88 & 6.26$\pm$0.17 & 3.37$\pm$0.27 & 8.25$\pm$0.42 & 8.20$\pm$0.22 & 4.50$\pm$0.58 & 11.65$\pm$0.82 & 8.58$\pm$0.20 & 5.17$\pm$0.56 & 13.42$\pm$0.68
 \\ 
 & FOGS & 3.04$\pm$0.24 & 2.03$\pm$0.13 & 3.78$\pm$0.54 & 6.12$\pm$0.18 & 3.19$\pm$0.10 & 8.34$\pm$0.56 & 7.91$\pm$0.17 & 4.37$\pm$0.12 & 11.16$\pm$0.43 & 8.31$\pm$0.18 & 4.87$\pm$0.20 & 12.78$\pm$0.43
\\
 & FEDFormer & 6.54$\pm$0.20 & 4.75$\pm$0.25 & 9.63$\pm$0.31 & 6.80$\pm$0.19 & 4.78$\pm$0.20 & 9.90$\pm$0.26 & \underline{7.59$\pm$0.14} & 4.28$\pm$0.16 & \underline{9.79$\pm$0.22} & 8.53$\pm$0.40 & 4.87$\pm$0.46 & 11.07$\pm$0.51
\\
 & NLinear & 3.05$\pm$0.52 & 1.87$\pm$0.46 & 3.65$\pm$1.15 & \underline{5.96$\pm$0.13} & 3.46$\pm$0.09 & \underline{7.02$\pm$0.27} & 7.74$\pm$0.17 & 4.31$\pm$0.09 & 11.99$\pm$0.32 & \underline{8.05$\pm$0.20} & \underline{4.62$\pm$0.13} & \underline{10.64$\pm$0.22}
\\
 & AdaRNN & 3.17$\pm$0.18 & 2.09$\pm$0.12 & 4.10$\pm$0.33 & 6.25$\pm$0.15 & 3.41$\pm$0.13 & 8.19$\pm$0.26 & 7.98$\pm$0.31 & 4.40$\pm$0.26 & 12.02$\pm$0.50 & 8.70$\pm$0.47 & 5.19$\pm$0.37 & 13.69$\pm$0.59
\\ 

 & ST-GFSL & 3.90$\pm$0.64 & 2.74$\pm$0.61 & 5.34$\pm$1.44 & 6.68$\pm$0.22 & 3.85$\pm$0.46 & 8.80$\pm$0.69 & 8.33$\pm$0.21 & 4.71$\pm$0.61 & 11.61$\pm$0.71 & 9.20$\pm$0.18 & 5.29$\pm$0.64 & 13.21$\pm$0.79
\\

 & STEP & 3.14$\pm$0.22 & 1.90$\pm$0.13 & 4.21$\pm$0.40 & 6.32$\pm$0.10 & 3.45$\pm$0.10 & 8.18$\pm$0.20 & 8.05$\pm$0.17 & 4.57$\pm$0.10 & 10.78$\pm$0.15 & 8.79$\pm$0.15 & 4.91$\pm$0.13 & 12.17$\pm$0.20
\\ 
 & TPB & \underline{3.01$\pm$0.09} & \underline{1.70$\pm$0.04} & \underline{3.51$\pm$0.08} & 5.98$\pm$0.09 & 3.23$\pm$0.09 & 7.59$\pm$0.27 & 7.70$\pm$0.15 & \underline{4.24$\pm$0.15} & 10.73$\pm$0.17 & 8.72$\pm$0.15 & 4.96$\pm$0.29 & 12.72$\pm$0.16
\\
 & MTPB & \cellcolor{lgray}{\textbf{2.52$\pm$0.04**}} & \cellcolor{lgray}{\textbf{1.59$\pm$0.05*}} & \cellcolor{lgray}{\textbf{3.24$\pm$0.10*}} & \cellcolor{lgray}{\textbf{5.44$\pm$0.05**}} & \cellcolor{lgray}{\textbf{2.96$\pm$0.13**}} & \cellcolor{lgray}{\textbf{6.84$\pm$0.15**}} & \cellcolor{lgray}{\textbf{6.41$\pm$0.11**}} & \cellcolor{lgray}{\textbf{3.53$\pm$0.13**}} & \cellcolor{lgray}{\textbf{8.29$\pm$0.28**}} & \cellcolor{lgray}{\textbf{6.90$\pm$0.08**}} & \cellcolor{lgray}{\textbf{3.89$\pm$0.13**}} & \cellcolor{lgray}{\textbf{8.98$\pm$0.15**}}
\\ 

\midrule
\midrule

\multirow{ 11}{*}{\rotatebox{90}{P,C,S$\rightarrow$METR-LA}}& HA & 11.809 & 7.575 & 22.848 & 11.809 & 7.575 & 22.848& 11.809 & 7.575 & 22.848& 11.809 & 7.575 & 22.848 
\\ 
 & DCRNN & 5.79$\pm$0.18 & 3.61$\pm$0.16 & 9.32$\pm$0.52 & 8.85$\pm$0.17 & 5.16$\pm$0.11 & 14.53$\pm$0.42 & 10.71$\pm$0.27 & 6.32$\pm$0.15 & 18.49$\pm$0.50 & 11.66$\pm$0.37 & 7.01$\pm$0.23 & 20.52$\pm$0.62
\\ 
 & GWN & 5.39$\pm$0.09 & 3.18$\pm$0.10 & 7.92$\pm$0.36 & 8.94$\pm$0.24 & 5.15$\pm$0.20 & 15.12$\pm$0.70 & 10.88$\pm$0.25 & 6.49$\pm$0.20 & 19.62$\pm$0.47 & 11.64$\pm$0.27 & 7.18$\pm$0.20 & 21.88$\pm$0.49
 \\ 
 & STFGNN & 5.35$\pm$0.20 & 3.16$\pm$0.20 & 7.98$\pm$0.68 & 8.92$\pm$0.31 & 5.12$\pm$0.21 & 14.83$\pm$0.56 & 10.62$\pm$0.29 & 6.40$\pm$0.27 & 19.54$\pm$0.51 & 11.47$\pm$0.24 & 7.18$\pm$0.29 & 22.15$\pm$0.59
 \\ 
 & DSTAGNN & 5.35$\pm$0.45 & 3.33$\pm$0.64 & 8.45$\pm$1.46 & 8.79$\pm$0.18 & 5.04$\pm$0.24 & 14.98$\pm$0.85 & 10.64$\pm$0.24 & 6.57$\pm$0.38 & 20.10$\pm$0.87 & 11.54$\pm$0.26 & 7.43$\pm$0.37 & 23.25$\pm$0.65
 \\ 
 & FOGS & 5.48$\pm$0.08 & 3.15$\pm$0.12 & 8.62$\pm$0.60 & \underline{8.69$\pm$0.18} & 5.00$\pm$0.11 & 14.76$\pm$0.65 & \underline{10.46$\pm$0.22} & \underline{6.28$\pm$0.17} & 19.76$\pm$0.75 & 11.34$\pm$0.21 & 6.97$\pm$0.28 & 22.30$\pm$0.69
 \\ 

 & FEDFormer & 10.22$\pm$0.28 & 6.48$\pm$0.33 & 18.57$\pm$0.68 & 10.01$\pm$0.20 & 6.60$\pm$0.35 & 18.38$\pm$0.65 & 10.96$\pm$0.15 & 6.61$\pm$0.28 & 18.62$\pm$0.79 & \underline{11.06$\pm$0.17} & \underline{6.40$\pm$0.35} & \underline{18.09$\pm$0.59}
\\
 & NLinear & 5.38$\pm$0.08 & 3.17$\pm$0.07 & 8.05$\pm$0.44 & 8.98$\pm$0.13 & \underline{4.92$\pm$0.14} & 14.96$\pm$0.55 & 10.69$\pm$0.13 & 6.41$\pm$0.08 & \underline{18.37$\pm$0.49} & 11.15$\pm$0.16 & 6.52$\pm$0.16 & 19.90$\pm$0.52
\\
 & AdaRNN & 5.33$\pm$0.43 & 3.15$\pm$0.71 & 9.22$\pm$1.27 & 8.75$\pm$0.17 & 5.18$\pm$0.24 & 15.64$\pm$0.74 & 10.69$\pm$0.27 & 6.69$\pm$0.29 & 19.53$\pm$0.86 & 11.78$\pm$0.27 & 7.61$\pm$0.38 & 23.19$\pm$0.51
\\ 
 & ST-GFSL & 5.80$\pm$0.54 & 3.93$\pm$0.94 & 9.04$\pm$1.12 & 9.13$\pm$0.42 & 5.91$\pm$1.13 & 15.68$\pm$1.37 & 10.87$\pm$0.27 & 7.28$\pm$0.80 & 20.11$\pm$0.97 & 11.78$\pm$0.15 & 8.03$\pm$0.58 & 22.59$\pm$0.60
 \\ 

 & STEP & 5.67$\pm$0.16 & 3.57$\pm$0.17 & 9.56$\pm$0.71 & 9.12$\pm$0.11 & 5.54$\pm$0.17 & 15.99$\pm$0.46 & 10.65$\pm$0.18 & 6.57$\pm$0.15 & 19.02$\pm$0.28 & 11.20$\pm$0.18 & 7.02$\pm$0.15 & 20.80$\pm$0.24

\\ 
 & TPB & \underline{5.27$\pm$0.13} & \underline{3.14$\pm$0.11} & \underline{7.87$\pm$0.35} & 8.88$\pm$0.21 & 5.03$\pm$0.15 & \underline{14.48$\pm$0.30} & 10.80$\pm$0.28 & 6.30$\pm$0.25 & 18.70$\pm$0.41 & 11.48$\pm$0.21 & 6.91$\pm$0.39 & 20.46$\pm$0.38
\\
 & MTPB & \cellcolor{lgray}{\textbf{4.92$\pm$0.05*}} & \cellcolor{lgray}{\textbf{2.90$\pm$0.05*}} & \cellcolor{lgray}{\textbf{7.37$\pm$0.17*}} & \cellcolor{lgray}{\textbf{8.04$\pm$0.16**}} & \cellcolor{lgray}{\textbf{4.70$\pm$0.07**}} & \cellcolor{lgray}{\textbf{14.46$\pm$0.25*}} & \cellcolor{lgray}{\textbf{9.96$\pm$0.12**}} & \cellcolor{lgray}{\textbf{5.38$\pm$0.10**}} & \cellcolor{lgray}{\textbf{16.70$\pm$0.33**}} & \cellcolor{lgray}{\textbf{10.31$\pm$0.20**}} & \cellcolor{lgray}{\textbf{5.71$\pm$0.10**}} & \cellcolor{lgray}{\textbf{17.70$\pm$0.46**}}
\\ 
\midrule
\midrule

\multirow{ 11}{*}{\rotatebox{90}{P,M,S$\rightarrow$Chengdu}}& HA & 6.860 & 5.253 & 22.497 & 6.860 & 5.253 & 22.497& 6.860 & 5.253 & 22.497& 6.860 & 5.253 & 22.497 \\ 
 & DCRNN & 3.51$\pm$0.06 & 2.46$\pm$0.04 & 10.55$\pm$0.29 & 4.82$\pm$0.08 & 3.40$\pm$0.06 & 14.95$\pm$0.33 & 5.73$\pm$0.14 & 4.15$\pm$0.09 & 18.25$\pm$0.47 & 6.38$\pm$0.12 & 4.72$\pm$0.08 & 20.63$\pm$0.64
\\ 
 & GWN & 3.32$\pm$0.09 & 2.43$\pm$0.07 & 9.97$\pm$0.35 & 4.82$\pm$0.08 & 3.39$\pm$0.06 & 14.79$\pm$0.46 & 5.81$\pm$0.12 & 4.21$\pm$0.09 & 18.51$\pm$0.66 & 6.51$\pm$0.12 & 4.85$\pm$0.09 & 21.08$\pm$0.65
 \\ 
 & STFGNN & 3.48$\pm$0.09 & 2.45$\pm$0.08 & 10.21$\pm$0.54 & 4.89$\pm$0.08 & 3.45$\pm$0.06 & 15.08$\pm$0.56 & 5.91$\pm$0.08 & 4.31$\pm$0.05 & 19.18$\pm$0.65 & 6.64$\pm$0.07 & 4.99$\pm$0.06 & 21.94$\pm$0.81
 \\ 
 & DSTAGNN & 3.47$\pm$0.13 & 2.44$\pm$0.15 & 10.38$\pm$0.53 & 4.99$\pm$0.21 & 3.56$\pm$0.25 & 15.69$\pm$0.90 & 6.01$\pm$0.24 & 4.39$\pm$0.21 & 19.62$\pm$1.19 & 6.71$\pm$0.26 & 5.04$\pm$0.23 & 22.22$\pm$1.53
 \\ 
 & FOGS & 3.37$\pm$0.11 & 2.37$\pm$0.09 & 9.90$\pm$0.58 & 4.76$\pm$0.08 & 3.35$\pm$0.07 & 14.79$\pm$0.55 & 5.75$\pm$0.13 & 4.18$\pm$0.12 & 18.64$\pm$0.86 & 6.34$\pm$0.13 & 4.71$\pm$0.14 & 20.98$\pm$1.03
 \\ 

 & FEDFormer & 4.33$\pm$0.09 & 2.81$\pm$0.13 & 12.55$\pm$0.28 & 4.28$\pm$0.08 & 2.83$\pm$0.11 & 12.56$\pm$0.23 & \underline{4.38$\pm$0.07} & \underline{2.95$\pm$0.12} & \underline{12.56$\pm$0.20} & \underline{4.30$\pm$0.09} & 3.31$\pm$0.11 & \underline{13.09$\pm$0.27}
\\
 & NLinear & 3.32$\pm$0.03 & 2.26$\pm$0.03 & 9.64$\pm$0.21 & \underline{4.23$\pm$0.07} & \underline{2.74$\pm$0.05} & \underline{12.09$\pm$0.28} & 4.50$\pm$0.04 & 2.97$\pm$0.03 & 13.15$\pm$0.08 & 4.77$\pm$0.06 & \underline{3.05$\pm$0.04} & 13.55$\pm$0.20
\\
 & AdaRNN & 3.37$\pm$0.11 & 2.33$\pm$0.10 & 10.16$\pm$0.53 & 4.92$\pm$0.09 & 3.49$\pm$0.06 & 14.79$\pm$0.52 & 5.83$\pm$0.14 & 4.16$\pm$0.13 & 19.39$\pm$0.78 & 6.25$\pm$0.14 & 5.10$\pm$0.16 & 22.29$\pm$0.98
 \\

 & ST-GFSL & 3.56$\pm$0.13 & 2.54$\pm$0.15 & 10.31$\pm$0.32 & 4.79$\pm$0.06 & 3.38$\pm$0.04 & 14.13$\pm$0.18 & 5.55$\pm$0.06 & 3.96$\pm$0.05 & 16.82$\pm$0.16 & 6.10$\pm$0.04 & 4.40$\pm$0.10 & 18.97$\pm$0.09
\\

 & STEP & 3.52$\pm$0.13 & 2.51$\pm$0.11 & 10.12$\pm$0.41 & 4.67$\pm$0.10 & 3.30$\pm$0.08 & 13.91$\pm$0.30 & 5.07$\pm$0.12 & 3.60$\pm$0.10 & 15.26$\pm$0.40 & 5.21$\pm$0.14 & 3.72$\pm$0.12 & 15.97$\pm$0.43

\\ 
 & TPB & \underline{3.20$\pm$0.05} & \underline{2.13$\pm$0.04} & \underline{9.27$\pm$0.27} & 4.52$\pm$0.07 & 3.18$\pm$0.05 & 13.76$\pm$0.20 & 4.99$\pm$0.11 & 3.42$\pm$0.12 & 15.12$\pm$0.02 & 5.12$\pm$0.14 & 4.00$\pm$0.21 & 16.05$\pm$0.11
\\

 & MTPB & \cellcolor{lgray}{\textbf{3.08$\pm$0.03**}} & \cellcolor{lgray}{\textbf{2.13$\pm$0.03**}} & \cellcolor{lgray}{\textbf{8.84$\pm$0.22**}} & \cellcolor{lgray}{\textbf{3.92$\pm$0.04**}} & \cellcolor{lgray}{\textbf{2.67$\pm$0.03**}} & \cellcolor{lgray}{\textbf{11.38$\pm$0.09**}} & \cellcolor{lgray}{\textbf{4.05$\pm$0.06**}} & \cellcolor{lgray}{\textbf{2.77$\pm$0.04**}} & \cellcolor{lgray}{\textbf{11.86$\pm$0.14**}} & \cellcolor{lgray}{\textbf{4.09$\pm$0.04**}} & \cellcolor{lgray}{\textbf{2.81$\pm$0.03**}} & \cellcolor{lgray}{\textbf{12.03$\pm$0.10**}}

\\
\midrule
\midrule

\multirow{ 11}{*}{\rotatebox{90}{P,M,C$\rightarrow$Shenzhen}} & HA & 5.491 & 3.924 & 15.521 & 5.491 & 3.924 & 15.521 & 5.491 & 3.924 & 15.521 & 5.491 & 3.924 & 15.521 \\ 
 & DCRNN & 3.02$\pm$0.11 & 2.08$\pm$0.08 & 8.27$\pm$0.33 & 4.10$\pm$0.07 & 2.78$\pm$0.06 & 11.22$\pm$0.21 & 4.74$\pm$0.08 & 3.27$\pm$0.07 & 13.07$\pm$0.23 & 5.11$\pm$0.08 & 3.58$\pm$0.07 & 14.30$\pm$0.26
\\ 
 & GWN & 2.93$\pm$0.08 & 2.02$\pm$0.08 & 7.99$\pm$0.18 & 4.21$\pm$0.06 & 2.86$\pm$0.06 & 11.38$\pm$0.15 & 5.04$\pm$0.07 & 3.45$\pm$0.07 & 13.85$\pm$0.28 & 5.55$\pm$0.09 & 3.89$\pm$0.06 & 15.56$\pm$0.40
\\ 
 & STFGNN & 3.04$\pm$0.05 & 2.09$\pm$0.06 & 8.21$\pm$0.23 & 4.28$\pm$0.05 & 2.90$\pm$0.05 & 11.68$\pm$0.27 & 5.08$\pm$0.05 & 3.50$\pm$0.04 & 14.14$\pm$0.36 & 5.62$\pm$0.05 & 3.98$\pm$0.06 & 15.88$\pm$0.53
 \\ 
 & DSTAGNN & 3.09$\pm$0.18 & 2.15$\pm$0.20 & 8.37$\pm$0.53 & 4.43$\pm$0.20 & 3.04$\pm$0.22 & 12.28$\pm$1.02 & 5.28$\pm$0.22 & 3.70$\pm$0.25 & 14.84$\pm$1.23 & 5.79$\pm$0.23 & 4.12$\pm$0.23 & 16.52$\pm$1.19
 \\ 
 & FOGS & 2.98$\pm$0.06 & 2.05$\pm$0.05 & 8.10$\pm$0.35 & 4.24$\pm$0.06 & 2.86$\pm$0.05 & 11.62$\pm$0.42 & 5.04$\pm$0.05 & 3.46$\pm$0.05 & 14.10$\pm$0.40 & 5.54$\pm$0.07 & 3.92$\pm$0.08 & 15.70$\pm$0.52
\\ 


 & FEDFormer & 3.77$\pm$0.10 & 2.81$\pm$0.11 & 10.43$\pm$0.41 & 3.84$\pm$0.11 & 2.66$\pm$0.16 & 10.11$\pm$0.36 & \underline{4.21$\pm$0.09} & 2.75$\pm$0.16 & \underline{10.36$\pm$0.38} & 4.65$\pm$0.09 & \underline{2.90$\pm$0.16} & \underline{11.19$\pm$0.35}
\\
 & NLinear & 2.97$\pm$0.03 & 1.96$\pm$0.03 & 7.62$\pm$0.11 & 3.81$\pm$0.03 & \underline{2.50$\pm$0.03} & \underline{10.04$\pm$0.08} & 4.47$\pm$0.04 & \underline{2.74$\pm$0.03} & 10.61$\pm$0.09 & \underline{4.43$\pm$0.04} & 3.01$\pm$0.04 & 11.58$\pm$0.13
\\
 & AdaRNN & 3.06$\pm$0.06 & 2.14$\pm$0.05 & 8.40$\pm$0.22 & 4.42$\pm$0.05 & 3.11$\pm$0.05 & 11.42$\pm$0.26 & 4.94$\pm$0.06 & 3.47$\pm$0.04 & 14.05$\pm$0.35 & 5.44$\pm$0.06 & 3.86$\pm$0.06 & 15.82$\pm$0.53
\\ 
 & ST-GFSL & 3.56$\pm$0.50 & 2.70$\pm$0.58 & 9.82$\pm$1.60 & 4.44$\pm$0.34 & 3.10$\pm$0.46 & 11.88$\pm$1.03 & 5.06$\pm$0.16 & 3.42$\pm$0.27 & 13.71$\pm$0.31 & 5.50$\pm$0.10 & 3.77$\pm$0.14 & 15.10$\pm$0.40
 \\ 

 & STEP & 3.08$\pm$0.11 & 2.19$\pm$0.10 & 8.28$\pm$0.34 & 4.07$\pm$0.04 & 2.69$\pm$0.05 & 11.06$\pm$0.10 & 4.75$\pm$0.14 & 2.97$\pm$0.04 & 11.97$\pm$0.14 & 4.79$\pm$0.05 & 3.30$\pm$0.16 & 12.78$\pm$0.21
\\ 
 & TPB & \underline{2.75$\pm$0.03} & \underline{1.91$\pm$0.03} & \underline{7.39$\pm$0.05} & \underline{3.80$\pm$0.03} & 2.54$\pm$0.03 & 10.27$\pm$0.06 & 4.51$\pm$0.04 & 3.04$\pm$0.01 & 12.11$\pm$0.06 & 4.93$\pm$0.09 & 3.35$\pm$0.04 & 13.10$\pm$0.12
\\

 & MTPB & \cellcolor{lgray}{\textbf{2.74$\pm$0.01**}} & \cellcolor{lgray}{\textbf{1.82$\pm$0.01**}} & \cellcolor{lgray}{\textbf{7.22$\pm$0.03**}} & \cellcolor{lgray}{\textbf{3.63$\pm$0.06**}} & \cellcolor{lgray}{\textbf{2.40$\pm$0.04**}} & \cellcolor{lgray}{\textbf{9.36$\pm$0.12**}} & \cellcolor{lgray}{\textbf{3.86$\pm$0.05**}} & \cellcolor{lgray}{\textbf{2.53$\pm$0.05**}} & \cellcolor{lgray}{\textbf{9.86$\pm$0.13**}} & \cellcolor{lgray}{\textbf{3.90$\pm$0.05**}} & \cellcolor{lgray}{\textbf{2.57$\pm$0.04**}} & \cellcolor{lgray}{\textbf{10.04$\pm$0.06**}}
 
\\ 

\bottomrule
\bottomrule
\end{tabular}
}
\vspace{-.3cm}
\end{table*}

\noindent{\textbf{Baselines:}}
We select 12 baselines of four types to evaluate the performance of MTPB on the few-shot forecasting task. 
These baselines include traditional methods, typical deep traffic forecasting methods, time series forecasting methods, and cross-city traffic forecasting methods.
We use the model structure hyper-parameters reported by the original papers of these models or frameworks. 
Note that the typical deep-learning traffic forecasting and time series forecasting methods are implemented in \textit{Reptile}, which is a meta-learning framework. 
For the hyperparameter of the meta-learning framework, we hold the same sufficient strategy for meta-learning hyperparameter searching in all methods including MTPB to guarantee fairness.
The learning rate of meta-training \& fine-tuning is set to 5e-4, the meta-training epoch is searched in range(5, 30, 5), the fine-tuning epoch is searched in range(50, 400, 50), update\_step is searched in range(2, 5, 1) (larger update\_step would lead to GPU memory limit exceeded).
We guarantee the fairness of the comparisons of different baselines.
\begin{itemize}[leftmargin=*]
    \item Traditional methods: \textbf{HA}, uses the average of previous periods as predictions.
    \item Typical deep traffic forecasting methods: \textbf{DCRNN}~\cite{li2017diffusion} uses diffusion techniques and RNN. \textbf{GWN}~\cite{wu2019graph} utilizes adaptive adjacency matrix and dilated causal convolution. \textbf{STFGNN}~\cite{li2021spatial} uses DTW distance to construct the temporal graph and gated mechanism. \textbf{DSTAGNN}~\cite{lan2022dstagnn} constructs spatial-temporal aware graph. \textbf{FOGS}~\cite{rao2022fogs} uses node2vec and predicts the first-order difference.
    \item Time series forecasting methods: \textbf{FEDFormer}~\cite{zhou2022fedformer} utilized a frequency enhanced Transformer.  \textbf{DLinear}~\cite{zeng2023transformers} uses linear layers with time decompositions.
    \item Cross-city traffic forecasting methods: \textbf{AdaRNN}~\cite{du2021adarnn} reduces the distribution mismatch. \textbf{ST-GFSL}~\cite{lu2022spatio} learns the metaknowledge of traffic nodes to generate the parameter of the models. \textbf{STEP}~\cite{shao2022pre} reconstructs the large-ratio masked traffic to do per-taining on source datasets. \textbf{TPB}~\cite{liu2023cross} is the base version of MTPB.
\end{itemize}
Notably, we focus on cross-city few-shot univariate forecasting, which stems from the need to address scenarios where the target city has minimal univariate data and additional modalities are unavailable.
Consequently, methods such as STrans-GAN~\cite{zhang2022strans}, CrossTReS~\cite{jin2022selective}, and TransGTR~\cite{jin2023transferable} would utilize multimodal data or are limited to one-to-one city transfer settings and thus they are not able to adapt here.

\noindent{\textbf{Overall Performance:}}
\begin{figure*}[!t]
    \centering
    \includegraphics[width=0.83\linewidth]{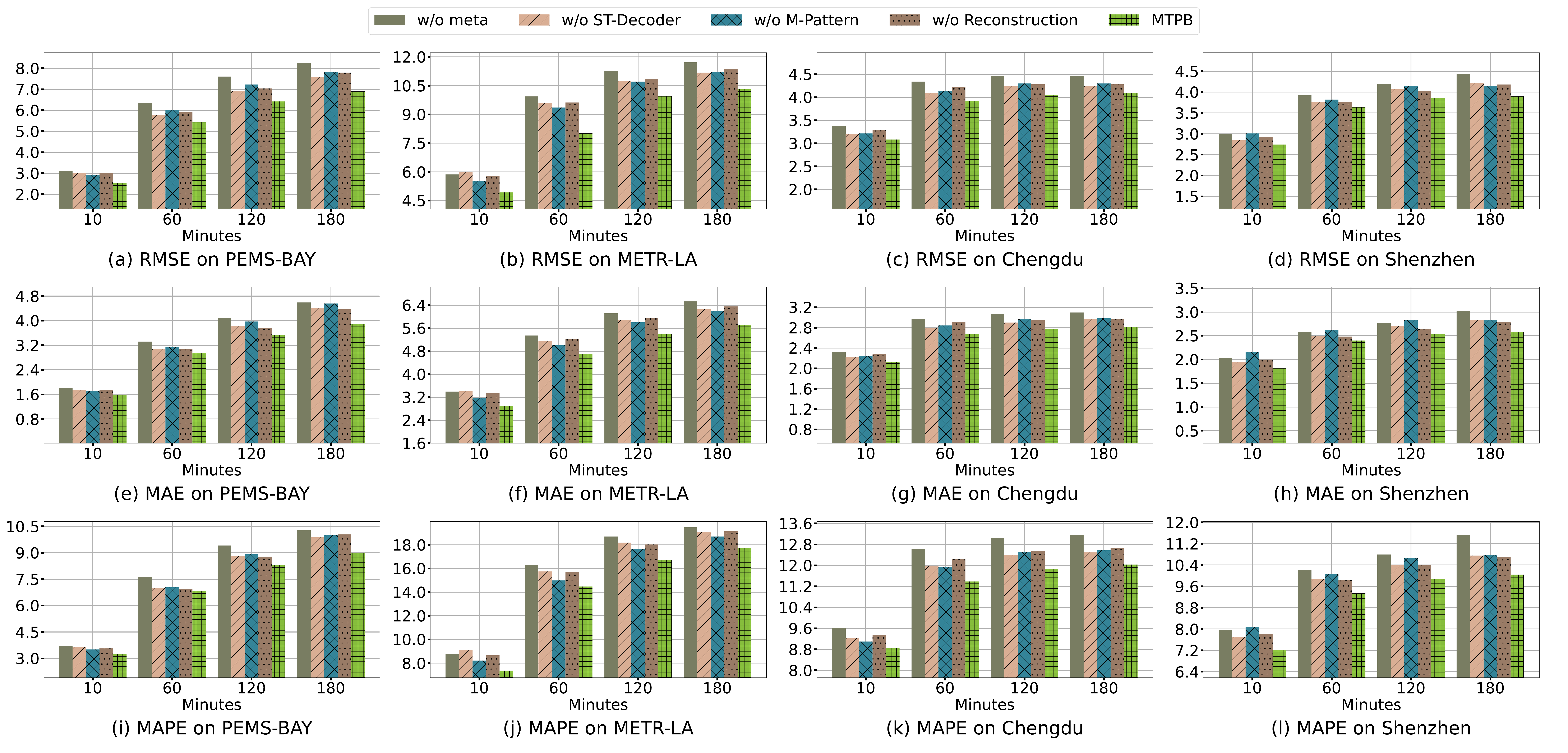}
    \caption{Ablation Study. We remove queried metaknowledge, ST-Decoder in Pretraining, Multi-scale Pattern, and Graph Reconstruction module separately and evaluate the performance.}
    \label{fig:4ablation}
\end{figure*}
The final performance results of both the baseline models and MTPB are presented in Table~\ref{tab:performance}. The analysis of the results reveals several key observations:
1) MTPB consistently outperforms the baseline models.
This success underscores the efficacy of the comprehensive multi-scale patterns, encompassing both short-term and long-term dynamics, in capturing resilient traffic patterns that generalize well across diverse cities.
2) Typical deep-learning traffic forecasting methods exhibit suboptimal performance, indicating that they lack the ability to capture robust traffic knowledge across cities and might aim to over-fit the training data of one city.
3) Time series forecasting methods have relatively strong performance in long-term forecasting, yet fall short in short-term accuracy.
4) Cross-city traffic forecasting methods predominantly prioritize short-term traffic information, often overlooking long-term patterns. Consequently, they exhibit commendable performance in short-term forecasting but fall short in long-term prediction.
5) TPB, the short-term pattern version of MTPB, demonstrates superior proficiency in short-term traffic forecasting. 
This indicates the effectiveness of the traffic pattern for cross-city transfer.
Furthermore, this emphasizes the importance of incorporating both short-term and long-term patterns, as evidenced by the heightened performance of MTPB in long-term forecasting.

\subsection{RQ2: Ablation Study}

In this section, we verify the effectiveness of each module of MTPB.
We remove or modify each module to get a variant of MTPB and do a grid search to get the best performance of these variants.
There are four types of variants of MTPB.
1) We remove the queried metaknowledge $\textbf{Z}$, denoted as \textbf{w/o meta}.
2) We remove ST-Decoder in Pre-training, denoted as \textbf{w/o ST-Decoder}.
3) We replace the multi-scale patterns with only short-term patterns, which is denoted as \textbf{w/o M-Pattern}.
4) We replace the reconstructed adjacency matrix with the pre-defined static adjacency matrix, which is denoted as \textbf{w/o Reconstruction}.
Results in Fig.~\ref{fig:4ablation} demonstrate that the performance declines without each module, with the absence of queried meta-knowledge $\textbf{Z}$ yielding the most significant impact.
These findings affirm the essential role of each module in MTPB, emphasizing their collective contribution to the model's efficacy in capturing diverse and complex traffic patterns across cities.

\begin{table}[!t]
\centering
\caption{The error of the Pre-training.
The average reconstruction error of the masked patches on test data is reported.
"$\backslash$" represents this dataset is not included in the source data.
"T" represents a transformer layer and "S" represents a GNN layer in ST-Decoder.}
\label{tab:sup_test}
\resizebox{0.96\linewidth}{!}{
\begin{tabular}{c|c c c|c c c|c c c|c c c}
\toprule
& \multicolumn{12}{c}{\textbf{Pretrain Datasets : METR-LA+Chendu+Shenzhen (PEMS-BAY not included)}}\\
\cline{2-13}
& \multicolumn{3}{c|}{\textbf{PEMS-BAY}} & \multicolumn{3}{c|}{\textbf{METR-LA}} & \multicolumn{3}{c|}{\textbf{Chengdu}} & \multicolumn{3}{c}{\textbf{Shenzhen}} \\

\cline{2-13}
& RMSE & MAE & MAPE & RMSE & MAE & MAPE & RMSE & MAE & MAPE & RMSE & MAE & MAPE \\
\midrule
\midrule

T  & $\backslash$ & $\backslash$ & $\backslash$ & 7.051 & 4.050 & 10.951 & 3.677 & 2.562 & 10.767 & 3.566 & 2.392 & 9.791 \\
T+S & $\backslash$ & $\backslash$ & $\backslash$ & 4.994 & 3.297 & 7.711 & 3.435 & 2.471 & 10.066 & 3.642 & 2.650 & 10.766 \\
T+S+T & $\backslash$ & $\backslash$ & $\backslash$ & \cellcolor{lgray}{3.870} & \cellcolor{lgray}{2.439} & \cellcolor{lgray}{5.548} & \cellcolor{lgray}{3.056} & \cellcolor{lgray}{2.166} & \cellcolor{lgray}{8.752} & \cellcolor{lgray}{2.877} & \cellcolor{lgray}{1.989} & \cellcolor{lgray}{8.112}   \\
\midrule
\midrule
& \multicolumn{12}{c}{\textbf{Pretrain Datasets : PEMS-BAY+Chendu+Shenzhen (METR-LA not included)}}\\
\cline{2-13}
& \multicolumn{3}{c|}{\textbf{PEMS-BAY}} & \multicolumn{3}{c|}{\textbf{METR-LA}} & \multicolumn{3}{c|}{\textbf{Chengdu}} & \multicolumn{3}{c}{\textbf{Shenzhen}} \\

\cline{2-13}
& RMSE & MAE & MAPE & RMSE & MAE & MAPE & RMSE & MAE & MAPE & RMSE & MAE & MAPE \\
\midrule
\midrule

T  & 4.444 & 2.389 & 5.150&$\backslash$ & $\backslash$ & $\backslash$  & 3.652 & 2.537 & 10.614 & 3.549 & 2.391 & 9.717 \\  
T+S & 3.838 & 2.414 & 4.899& $\backslash$ & $\backslash$ & $\backslash$  & 3.470 & 2.490 & 10.346 & 3.783 & 2.748 & 11.185 \\  
T+S+T & \cellcolor{lgray}{2.874} & \cellcolor{lgray}{1.779} & \cellcolor{lgray}{3.487} & $\backslash$ & $\backslash$ & $\backslash$ &  \cellcolor{lgray}{3.195} & \cellcolor{lgray}{2.233} & \cellcolor{lgray}{9.195} & \cellcolor{lgray}{3.035} & \cellcolor{lgray}{2.066} & \cellcolor{lgray}{8.396} \\ 

\midrule
\midrule
& \multicolumn{12}{c}{\textbf{Pretrain Datasets : PEMS-BAY+METR-LA+Shenzhen (Chengdu not included)}}\\
\cline{2-13}
& \multicolumn{3}{c|}{\textbf{PEMS-BAY}} & \multicolumn{3}{c|}{\textbf{METR-LA}} & \multicolumn{3}{c|}{\textbf{Chengdu}} & \multicolumn{3}{c}{\textbf{Shenzhen}} \\

\cline{2-13}
& RMSE & MAE & MAPE & RMSE & MAE & MAPE & RMSE & MAE & MAPE & RMSE & MAE & MAPE \\
\midrule
\midrule

T  & 4.620 & 2.423 & 5.560 &  7.035 & 3.988 & 10.940 &  $\backslash$ & $\backslash$ & $\backslash$ & 3.520 & 2.351 & 9.679 \\ 
T+S & 4.064 & 2.565 & 5.259 & 5.330 & 3.541 & 8.255 & $\backslash$ & $\backslash$ & $\backslash$ & 3.767 & 2.727 & 11.204 \\ 
T+S+T & \cellcolor{lgray}{3.246} & \cellcolor{lgray}{1.986} & \cellcolor{lgray}{3.936} & \cellcolor{lgray}{4.564} & \cellcolor{lgray}{2.827} & \cellcolor{lgray}{6.533} & $\backslash$ & $\backslash$ & $\backslash$ & \cellcolor{lgray}{3.096} & \cellcolor{lgray}{2.132} & \cellcolor{lgray}{8.613} \\   

\midrule
\midrule
& \multicolumn{12}{c}{\textbf{Pretrain Datasets : PEMS-BAY+METR-LA+Chengdu (Shenzhen not included)}}\\
\cline{2-13}
& \multicolumn{3}{c|}{\textbf{PEMS-BAY}} & \multicolumn{3}{c|}{\textbf{METR-LA}} & \multicolumn{3}{c|}{\textbf{Chengdu}} & \multicolumn{3}{c}{\textbf{Shenzhen}} \\

\cline{2-13}
& RMSE & MAE & MAPE & RMSE & MAE & MAPE & RMSE & MAE & MAPE & RMSE & MAE & MAPE \\
\midrule
\midrule

T  & 4.517 & 2.410 & 5.382 & 7.108 & 4.018 & 10.937 &  3.704 & 2.584 & 10.850 & $\backslash$ & $\backslash$ & $\backslash$  \\
T+S &  4.066 & 2.557 & 5.228 & 5.277 & 3.499 & 8.153  & 3.532 & 2.521 & 10.393 & $\backslash$ & $\backslash$ & $\backslash$ \\
T+S+T & \cellcolor{lgray}{2.920} & \cellcolor{lgray}{1.836} & \cellcolor{lgray}{3.597} & \cellcolor{lgray}{4.159} & \cellcolor{lgray}{ 2.645} & \cellcolor{lgray}{5.969} & \cellcolor{lgray}{3.286} & \cellcolor{lgray}{2.331} & \cellcolor{lgray}{9.543} &  $\backslash$ & $\backslash$ & $\backslash$ \\

\bottomrule
\bottomrule

\end{tabular}
}
\vspace{-.3cm}
\end{table}

\begin{figure}[!t]
    \centering
    \includegraphics[width=0.91\linewidth]{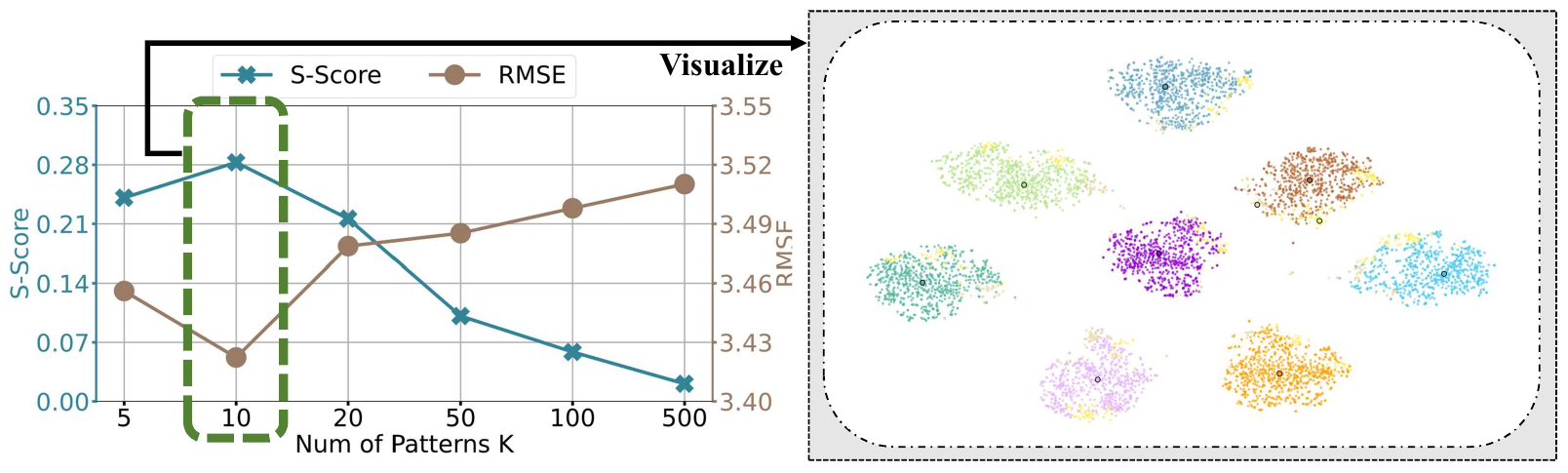}
    \caption{The performance and Silhouette score of traffic patterns w.r.t. different \textit{K} in \textit{Shenzhen}.}
    \label{fig:10sscore}
    \vspace{-.4cm}
\end{figure}

\subsection{RQ3: Pretraining}

\begin{figure*}[!t]
    \centering
    \includegraphics[width=0.86\linewidth]{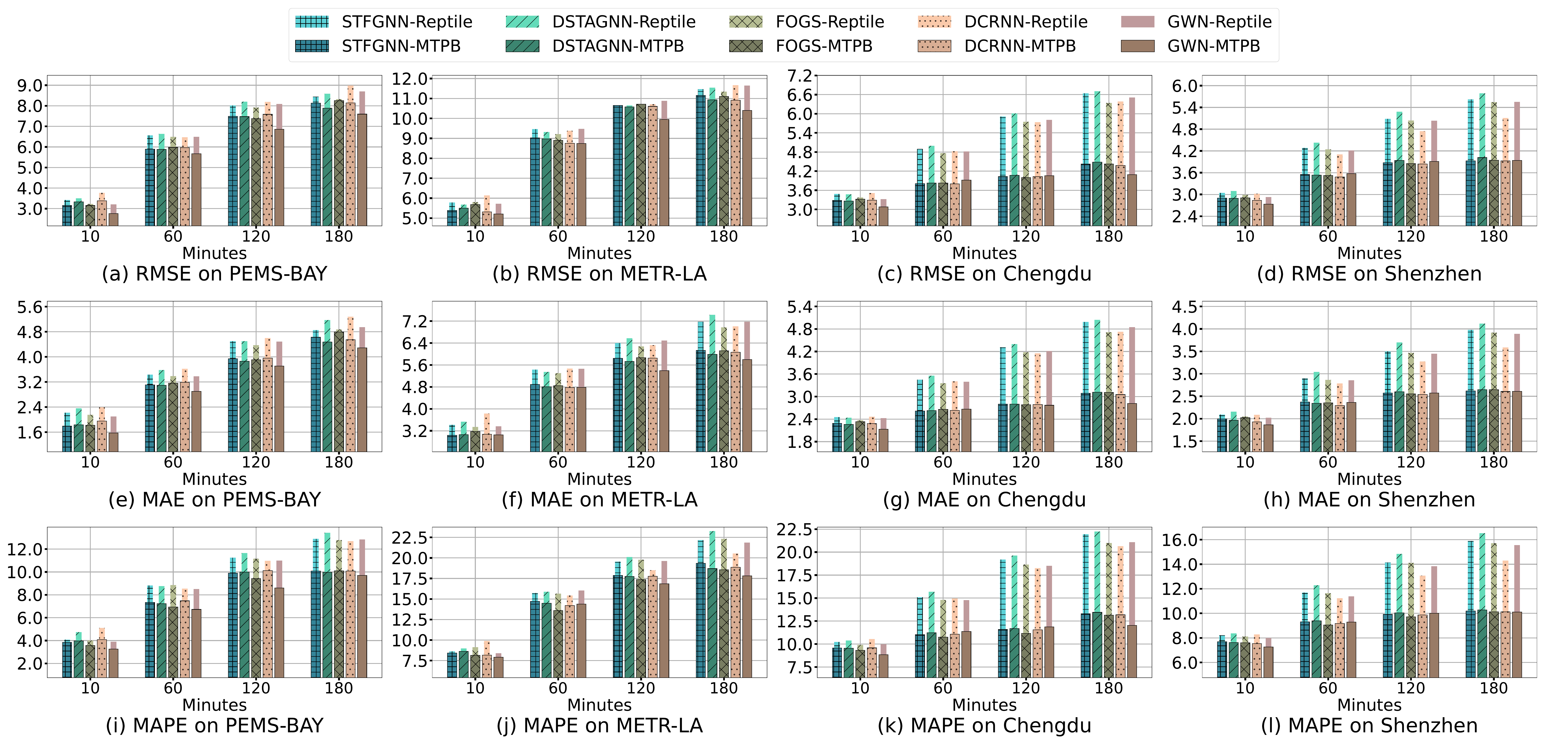}
    \caption{The performance of different STmodels trained in meta-learning framework Reptile and trained in our framework MTPB.}
    \label{fig:5addMTPB}
    \vspace{-.4cm}
\end{figure*}

\begin{figure}[!t]
    \centering
    \includegraphics[width=1\linewidth]{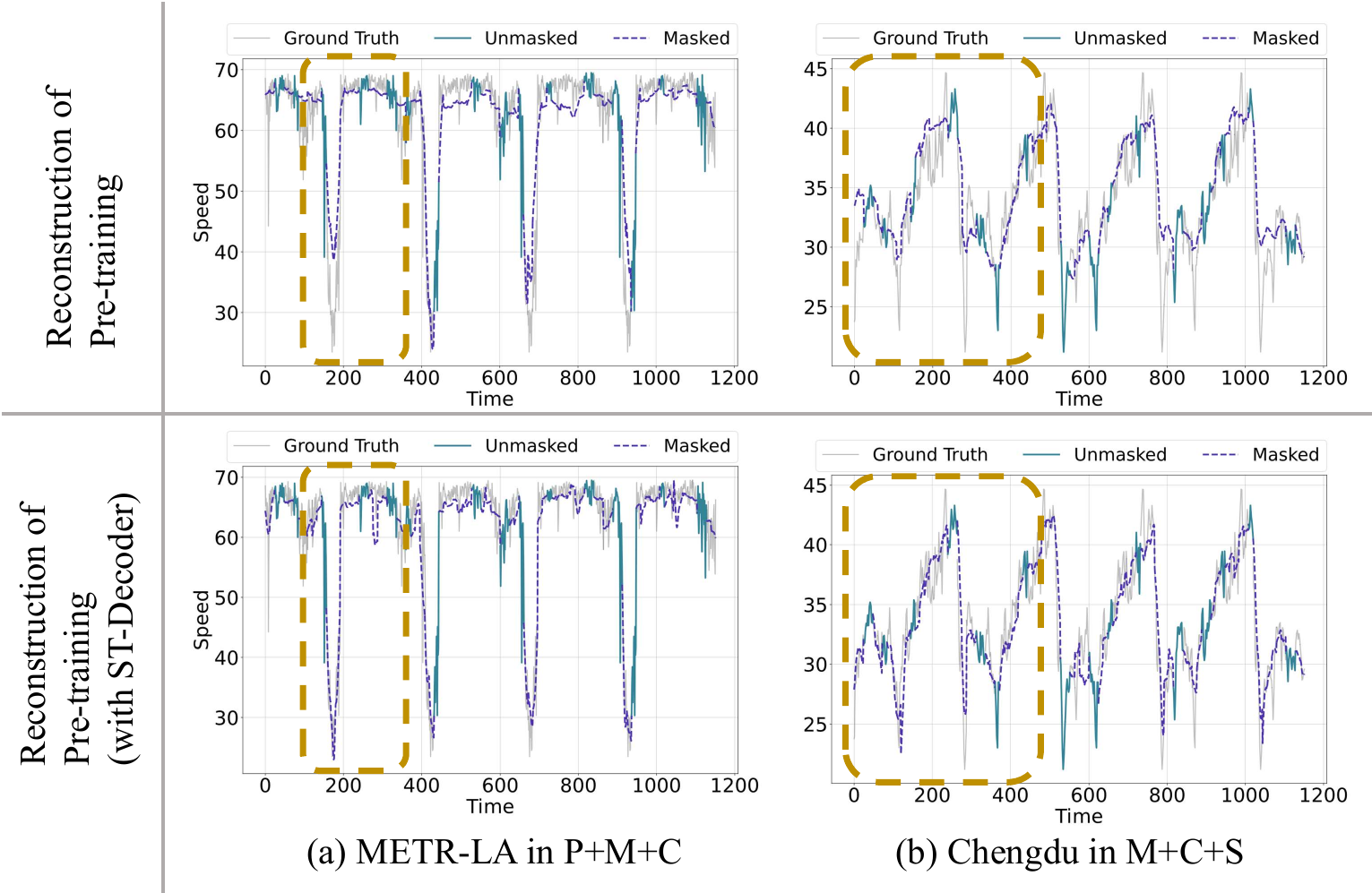}
    \caption{The reconstruction visualization of Pre-training. The purple dashed line is the reconstructed traffic. This figure shows that with the ST-Decoder, the reconstructed traffic is more accurate.}
    \label{fig:vis_pretrain}
    \vspace{-.1cm}
\end{figure}

This section investigates whether the Pre-training module effectively captures the robust knowledge from source cities.
By evaluating the reconstruction error of pre-trained models of each setting in RQ1, we scrutinize the Pretrain module's ability to capture nuanced traffic dynamics. 
Moreover, we evaluate the effectiveness of the ST-Decoder by sequentially adding spatial (S) and temporal (T) layers onto the fundamental transformer layer.
The result is shown in Table~\ref{tab:sup_test}.
From the table, we can see that
1) The reconstruction error in Table~\ref{tab:sup_test} of all settings is relatively low considering the high masking ratio (75\%), which indicates that robust traffic dynamics are captured.
2) By adding the spatial layer and temporal layer, the error declines drastically, which indicates that the additional spatial and temporal layer help better capture the robust traffic dynamics.
3) Furthermore, we visualize the reconstruction visualization in Fig.~\ref{fig:vis_pretrain}, which
shows the pre-trained model reconstructs the trend of the traffic successfully and ST-Decoder further enhances the reconstruction process.

\subsection{RQ4: Pattern Clustering: Selecet \textit{K}}

The clustering parameter, denoted as K, plays a crucial role in determining the quality of the generated traffic pattern bank during the pattern generation phase. To quantitatively assess the quality of the traffic pattern bank for different values of K, we calculate the Silhouette score~\cite{rousseeuw1987silhouettes} for the traffic pattern bank using the Shenzhen dataset, as illustrated in Fig.~\ref{fig:10sscore}.
Our observations are as follows:
1) When K is small (around 10), the Silhouette score is high, indicating good clustering quality, and the corresponding root mean square error (RMSE) is low.
2) As K increases, the Silhouette score decreases significantly, while the RMSE increases accordingly.
Based on these findings, we conclude that the Silhouette score is an effective criterion for evaluating the quality of the traffic pattern bank given a specific \textit{K}, and using a traffic pattern bank with a high Silhouette score in MTBP could lead to a small error.
We also show the TSNE visualization of the pattern bank with $K=10$, which demonstrates the successful capture of different traffic patterns through clustering.

\subsection{RQ5: Graph Reconstruction}

\begin{figure}[!t]
    \centering
    \includegraphics[width=.9\linewidth]{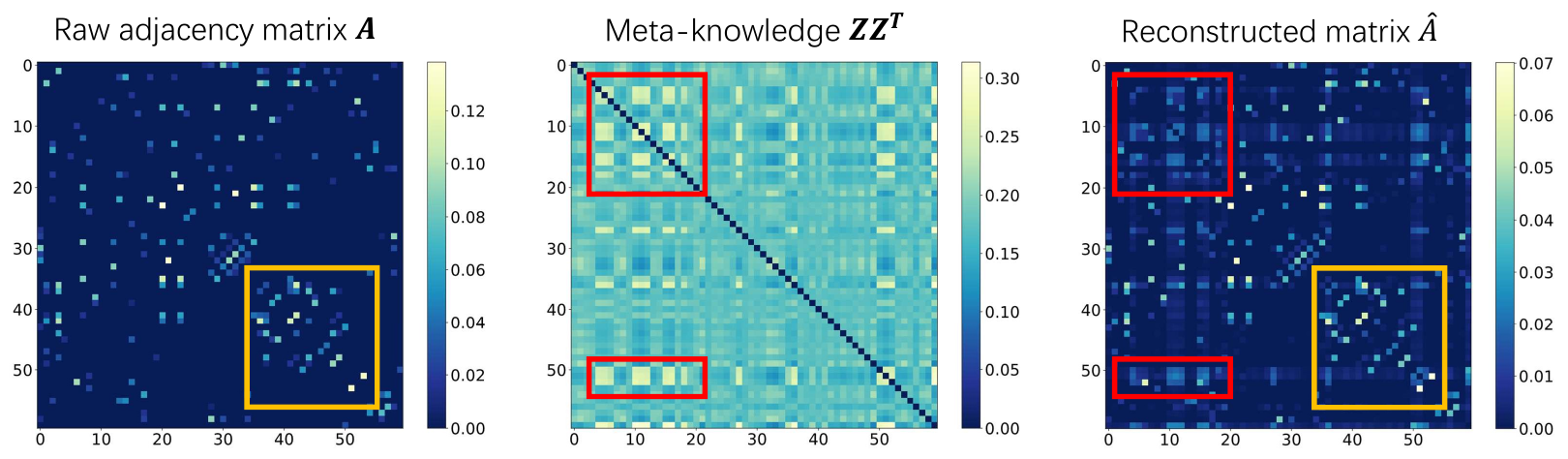}
    \caption{Visualization of the reconstructed adjacency matrix.}
    \label{fig:6recons}
\end{figure}

In this section, we explore the efficacy of the reconstruction module within the Pattern Aggregation component.
Visualizations of the raw adjacency matrix $\mathbf{A}$, the inner product of meta-knowledge $\mathbf{ZZ^T}$, and the reconstructed matrix $\mathbf{\hat{A}}$ have been generated.
For detailed insights, we remove the diagonal values and specifically visualize the submatrix nodes $40\sim 100$ in \textit{METR-LA} as depicted in Fig.~\ref{fig:6recons}.
The visualization reveals that both the patterns from the raw matrix (right rectangle) and the meta-knowledge (left rectangles) are preserved in the reconstructed matrix. 
This observation signifies the effectiveness of the self-expressive reconstruction method.

\subsection{RQ6: Differetn STmodel}

MTPB is a framework for few-shot traffic forecasting that different STmodels can plug into this framework.
To further verify the effectiveness of MTPB on the few-shot setting, we integrate several advanced STmodels into the MTPB framework and check the performance enhancement.
Fig.~\ref{fig:5addMTPB} shows the RMSE results of the aforementioned methods on four datasets of multi-step forecasting, from which we have the following observations:
(1) When integrated into the MTPB framework, all STmodels exhibit remarkably enhanced performance.
(2) The performance enhancement is more pronounced for the mid-term and long-term future steps.
(3) Simple STmodels such as Graph Wavenet have shown more significant performance improvement. 
This illustrates the efficacy of the multi-scale traffic pattern bank in enhancing the performance of STmodels. Notably, simpler STmodels exhibit greater flexibility, and the meta-knowledge readily adapts to these models with ease.

\subsection{RQ7: Efficiency}

\begin{figure}[!t]
    \centering
    \includegraphics[width=0.83\linewidth]{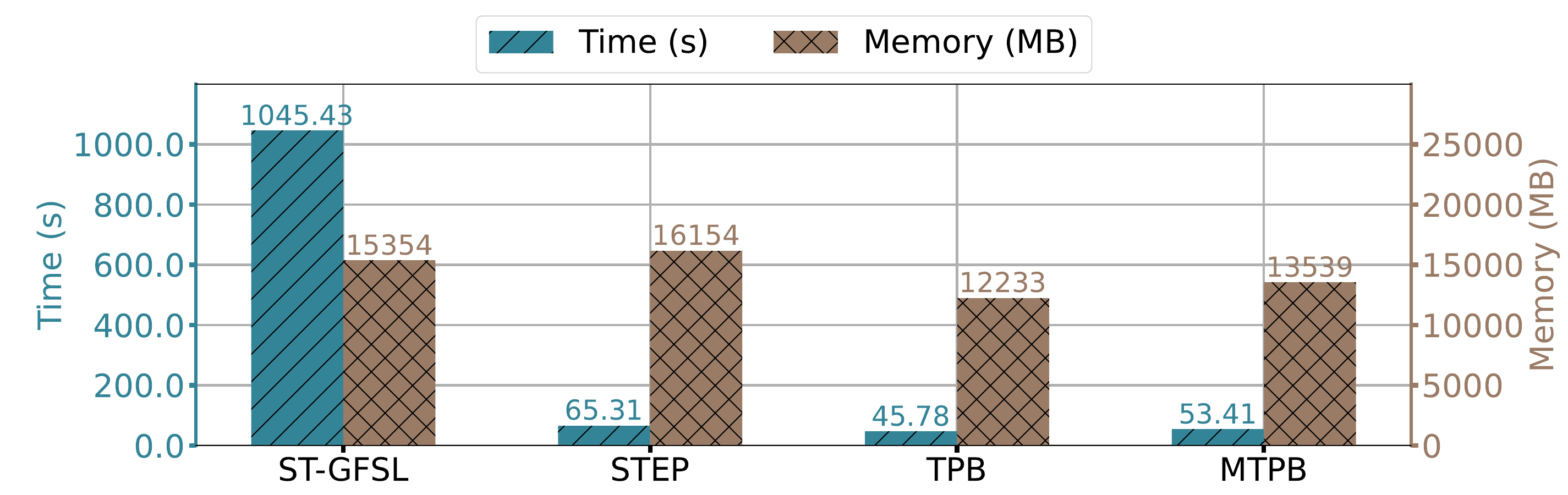}
    \caption{The efficiency comparison.}
    \label{fig:7efficiency}
    \vspace{-.2cm}
\end{figure}

In this section, we assess the efficiency of MTPB in comparison to various methods designed for cross-city traffic forecasting. Experiments are performed on \textit{METR-LA} using a single NVIDIA GTX3090 graphics card with 24GB memory. The total running time and GPU memory usage for a single training instance are reported in Fig.~\ref{fig:7efficiency}.
ST-GFSL exhibits an inefficiency, requiring over ten times the time of other baselines. 
Conversely, the remaining three methods demonstrate comparable acceptable running times and memory consumption, establishing their viability for application in the cross-city scenario.

\section{Conclusion}
In this work, we propose a novel cross-city few-shot traffic forecasting framework named MTPB.
We demonstrate that the traffic pattern is similar across cities.
To capture the similarity, we aim to generate a Multi-scale Traffic Pattern Bank to help downstream cross-city traffic forecasting.
Experiments on real-world traffic datasets demonstrate the superiority over state-of-the-art methods of the MTPB framework.
In the future, we will investigate the idea of the pattern bank of other types of time-series data such as financial and power-consuming data in the few-shot setting.


\ifCLASSOPTIONcompsoc
  \section*{Acknowledgments}
\else
  \section*{Acknowledgment}
\fi

This work was sponsored by National Key Research and Development Program of China under Grant No.2022YFB3904204, National Natural Science Foundation of China under Grant No.62102246, No.62272301, No.62176243, and Provincial Key Research and Development Program of Zhejiang under Grant No.2021C01034.
This paper is an extended version of \cite{liu2023cross} which has been published in the 32nd ACM International Conference on Information and Knowledge Management.
\bibliographystyle{IEEEtran}
\bibliography{ref}

\begin{thebibliography}{10}
\providecommand{\url}[1]{#1}
\csname url@samestyle\endcsname
\providecommand{\newblock}{\relax}
\providecommand{\bibinfo}[2]{#2}
\providecommand{\BIBentrySTDinterwordspacing}{\spaceskip=0pt\relax}
\providecommand{\BIBentryALTinterwordstretchfactor}{4}
\providecommand{\BIBentryALTinterwordspacing}{\spaceskip=\fontdimen2\font plus
\BIBentryALTinterwordstretchfactor\fontdimen3\font minus \fontdimen4\font\relax}
\providecommand{\BIBforeignlanguage}[2]{{%
\expandafter\ifx\csname l@#1\endcsname\relax
\typeout{** WARNING: IEEEtran.bst: No hyphenation pattern has been}%
\typeout{** loaded for the language `#1'. Using the pattern for}%
\typeout{** the default language instead.}%
\else
\language=\csname l@#1\endcsname
\fi
#2}}
\providecommand{\BIBdecl}{\relax}
\BIBdecl

\bibitem{li2019traffic}
J.~Li, D.~Fu, Q.~Yuan, H.~Zhang, K.~Chen, S.~Yang, and F.~Yang, ``A traffic prediction enabled double rewarded value iteration network for route planning,'' \emph{IEEE Transactions on Vehicular Technology}, vol.~68, no.~5, pp. 4170--4181, 2019.

\bibitem{wei2018intellilight}
H.~Wei, G.~Zheng, H.~Yao, and Z.~Li, ``Intellilight: A reinforcement learning approach for intelligent traffic light control,'' in \emph{Proceedings of the 24th ACM SIGKDD International Conference on Knowledge Discovery \& Data Mining}, 2018, pp. 2496--2505.

\bibitem{wang2022ctrl}
Y.~Wang, H.~Jin, and G.~Zheng, ``Ctrl: Cooperative traffic tolling via reinforcement learning,'' in \emph{Proceedings of the 31st ACM International Conference on Information \& Knowledge Management}, 2022, pp. 3545--3554.

\bibitem{li2017diffusion}
Y.~Li, R.~Yu, C.~Shahabi, and Y.~Liu, ``Diffusion convolutional recurrent neural network: Data-driven traffic forecasting,'' \emph{arXiv preprint arXiv:1707.01926}, 2017.

\bibitem{Didi}
G.~i. Didi, ``Didi chuxing data,'' \url{https://gaia.didichuxing.com}.

\bibitem{wang2018cross}
L.~Wang, X.~Geng, X.~Ma, F.~Liu, and Q.~Yang, ``Cross-city transfer learning for deep spatio-temporal prediction,'' \emph{arXiv preprint arXiv:1802.00386}, 2018.

\bibitem{jin2022selective}
Y.~Jin, K.~Chen, and Q.~Yang, ``Selective cross-city transfer learning for traffic prediction via source city region re-weighting,'' in \emph{Proceedings of the 28th ACM SIGKDD Conference on Knowledge Discovery and Data Mining}, 2022, pp. 731--741.

\bibitem{yao2019learning}
H.~Yao, Y.~Liu, Y.~Wei, X.~Tang, and Z.~Li, ``Learning from multiple cities: A meta-learning approach for spatial-temporal prediction,'' in \emph{The World Wide Web Conference}, 2019, pp. 2181--2191.

\bibitem{pan2019urban}
Z.~Pan, Y.~Liang, W.~Wang, Y.~Yu, Y.~Zheng, and J.~Zhang, ``Urban traffic prediction from spatio-temporal data using deep meta learning,'' in \emph{Proceedings of the 25th ACM SIGKDD international conference on knowledge discovery \& data mining}, 2019, pp. 1720--1730.

\bibitem{lu2022spatio}
B.~Lu, X.~Gan, W.~Zhang, H.~Yao, L.~Fu, and X.~Wang, ``Spatio-temporal graph few-shot learning with cross-city knowledge transfer,'' \emph{arXiv preprint arXiv:2205.13947}, 2022.

\bibitem{jin2023transferable}
Y.~Jin, K.~Chen, and Q.~Yang, ``Transferable graph structure learning for graph-based traffic forecasting across cities,'' in \emph{Proceedings of the 29th ACM SIGKDD Conference on Knowledge Discovery and Data Mining}, 2023, pp. 1032--1043.

\bibitem{he2022masked}
K.~He, X.~Chen, S.~Xie, Y.~Li, P.~Doll{\'a}r, and R.~Girshick, ``Masked autoencoders are scalable vision learners,'' in \emph{Proceedings of the IEEE/CVF Conference on Computer Vision and Pattern Recognition}, 2022, pp. 16\,000--16\,009.

\bibitem{kang2022fine}
Z.~Kang, Z.~Liu, S.~Pan, and L.~Tian, ``Fine-grained attributed graph clustering,'' in \emph{Proceedings of the 2022 SIAM International Conference on Data Mining (SDM)}.\hskip 1em plus 0.5em minus 0.4em\relax SIAM, 2022, pp. 370--378.

\bibitem{xu2019scaled}
J.~Xu, M.~Yu, L.~Shao, W.~Zuo, D.~Meng, L.~Zhang, and D.~Zhang, ``Scaled simplex representation for subspace clustering,'' \emph{IEEE Transactions on Cybernetics}, vol.~51, no.~3, pp. 1493--1505, 2019.

\bibitem{nichol2018first}
A.~Nichol, J.~Achiam, and J.~Schulman, ``On first-order meta-learning algorithms,'' \emph{arXiv preprint arXiv:1803.02999}, 2018.

\bibitem{nikravesh2016mobile}
A.~Y. Nikravesh, S.~A. Ajila, C.-H. Lung, and W.~Ding, ``Mobile network traffic prediction using mlp, mlpwd, and svm,'' in \emph{2016 IEEE International Congress on Big Data (BigData Congress)}.\hskip 1em plus 0.5em minus 0.4em\relax IEEE, 2016, pp. 402--409.

\bibitem{akagi2018fast}
Y.~Akagi, T.~Nishimura, T.~Kurashima, and H.~Toda, ``A fast and accurate method for estimating people flow from spatiotemporal population data.'' in \emph{IJCAI}, 2018, pp. 3293--3300.

\bibitem{liang2022cblab}
C.~Liang, Z.~Huang, Y.~Liu, Z.~Liu, G.~Zheng, H.~Shi, Y.~Du, F.~Li, and Z.~Li, ``Cblab: Scalable traffic simulation with enriched data supporting,'' \emph{arXiv preprint arXiv:2210.00896}, 2022.

\bibitem{yu2017spatio}
B.~Yu, H.~Yin, and Z.~Zhu, ``Spatio-temporal graph convolutional networks: A deep learning framework for traffic forecasting,'' \emph{arXiv preprint arXiv:1709.04875}, 2017.

\bibitem{yao2019revisiting}
H.~Yao, X.~Tang, H.~Wei, G.~Zheng, and Z.~Li, ``Revisiting spatial-temporal similarity: A deep learning framework for traffic prediction,'' 2018.

\bibitem{li2021spatial}
M.~Li and Z.~Zhu, ``Spatial-temporal fusion graph neural networks for traffic flow forecasting,'' in \emph{Proceedings of the AAAI conference on artificial intelligence}, vol.~35, no.~5, 2021, pp. 4189--4196.

\bibitem{gupta2023frigate}
M.~Gupta, H.~Kodamana, and S.~Ranu, ``Frigate: Frugal spatio-temporal forecasting on road networks,'' \emph{arXiv preprint arXiv:2306.08277}, 2023.

\bibitem{ma2023rethinking}
Q.~Ma, Z.~Zhang, X.~Zhao, H.~Li, H.~Zhao, Y.~Wang, Z.~Liu, and W.~Wang, ``Rethinking sensors modeling: Hierarchical information enhanced traffic forecasting,'' in \emph{Proceedings of the 32nd ACM International Conference on Information and Knowledge Management}, 2023, pp. 1756--1765.

\bibitem{bai2020adaptive}
L.~Bai, L.~Yao, C.~Li, X.~Wang, and C.~Wang, ``Adaptive graph convolutional recurrent network for traffic forecasting,'' \emph{arXiv preprint arXiv:2007.02842}, 2020.

\bibitem{wu2019graph}
Z.~Wu, S.~Pan, G.~Long, J.~Jiang, and C.~Zhang, ``Graph wavenet for deep spatial-temporal graph modeling,'' \emph{arXiv preprint arXiv:1906.00121}, 2019.

\bibitem{zheng2020gman}
C.~Zheng, X.~Fan, C.~Wang, and J.~Qi, ``Gman: A graph multi-attention network for traffic prediction,'' in \emph{Proceedings of the AAAI conference on artificial intelligence}, vol.~34, no.~01, 2020, pp. 1234--1241.

\bibitem{shao2022decoupled}
Z.~Shao, Z.~Zhang, W.~Wei, F.~Wang, Y.~Xu, X.~Cao, and C.~S. Jensen, ``Decoupled dynamic spatial-temporal graph neural network for traffic forecasting,'' \emph{arXiv preprint arXiv:2206.09112}, 2022.

\bibitem{cirstea2022towards}
R.-G. Cirstea, B.~Yang, C.~Guo, T.~Kieu, and S.~Pan, ``Towards spatio-temporal aware traffic time series forecasting,'' in \emph{2022 IEEE 38th International Conference on Data Engineering (ICDE)}.\hskip 1em plus 0.5em minus 0.4em\relax IEEE, 2022, pp. 2900--2913.

\bibitem{lan2022dstagnn}
S.~Lan, Y.~Ma, W.~Huang, W.~Wang, H.~Yang, and P.~Li, ``Dstagnn: Dynamic spatial-temporal aware graph neural network for traffic flow forecasting,'' in \emph{International Conference on Machine Learning}.\hskip 1em plus 0.5em minus 0.4em\relax PMLR, 2022, pp. 11\,906--11\,917.

\bibitem{jiang2023enhancing}
J.~Jiang, B.~Wu, L.~Chen, K.~Zhang, and S.~Kim, ``Enhancing the robustness via adversarial learning and joint spatial-temporal embeddings in traffic forecasting,'' in \emph{Proceedings of the 32nd ACM International Conference on Information and Knowledge Management}, 2023, pp. 987--996.

\bibitem{zhang2023mask}
X.~Zhang, Y.~Gong, X.~Zhang, X.~Wu, C.~Zhang, and X.~Dong, ``Mask-and contrast-enhanced spatio-temporal learning for urban flow prediction,'' in \emph{Proceedings of the 32nd ACM International Conference on Information and Knowledge Management}, 2023, pp. 3298--3307.

\bibitem{duan2023localised}
W.~Duan, X.~He, Z.~Zhou, L.~Thiele, and H.~Rao, ``Localised adaptive spatial-temporal graph neural network,'' \emph{arXiv preprint arXiv:2306.06930}, 2023.

\bibitem{wu2020connecting}
Z.~Wu, S.~Pan, G.~Long, J.~Jiang, X.~Chang, and C.~Zhang, ``Connecting the dots: Multivariate time series forecasting with graph neural networks,'' in \emph{Proceedings of the 26th ACM SIGKDD international conference on knowledge discovery \& data mining}, 2020, pp. 753--763.

\bibitem{han2021dynamic}
L.~Han, B.~Du, L.~Sun, Y.~Fu, Y.~Lv, and H.~Xiong, ``Dynamic and multi-faceted spatio-temporal deep learning for traffic speed forecasting,'' in \emph{Proceedings of the 27th ACM SIGKDD Conference on Knowledge Discovery \& Data Mining}, 2021, pp. 547--555.

\bibitem{choi2022graph}
J.~Choi, H.~Choi, J.~Hwang, and N.~Park, ``Graph neural controlled differential equations for traffic forecasting,'' in \emph{Proceedings of the AAAI Conference on Artificial Intelligence}, vol.~36, no.~6, 2022, pp. 6367--6374.

\bibitem{ji2022stden}
J.~Ji, J.~Wang, Z.~Jiang, J.~Jiang, and H.~Zhang, ``{STDEN}: Towards physics-guided neural networks for traffic flow prediction,'' in \emph{Proceedings of the AAAI Conference on Artificial Intelligence}, vol.~36, no.~4, 2022, pp. 4048--4056.

\bibitem{diao2019dynamic}
Z.~Diao, X.~Wang, D.~Zhang, Y.~Liu, K.~Xie, and S.~He, ``Dynamic spatial-temporal graph convolutional neural networks for traffic forecasting,'' in \emph{Proceedings of the AAAI conference on artificial intelligence}, vol.~33, no.~01, 2019, pp. 890--897.

\bibitem{cao2020spectral}
D.~Cao, Y.~Wang, J.~Duan, C.~Zhang, X.~Zhu, C.~Huang, Y.~Tong, B.~Xu, J.~Bai, J.~Tong \emph{et~al.}, ``Spectral temporal graph neural network for multivariate time-series forecasting,'' \emph{Advances in neural information processing systems}, vol.~33, pp. 17\,766--17\,778, 2020.

\bibitem{lee2021learning}
H.~Lee, S.~Jin, H.~Chu, H.~Lim, and S.~Ko, ``Learning to remember patterns: Pattern matching memory networks for traffic forecasting,'' \emph{arXiv preprint arXiv:2110.10380}, 2021.

\bibitem{shao2022pre}
Z.~Shao, Z.~Zhang, F.~Wang, and Y.~Xu, ``Pre-training enhanced spatial-temporal graph neural network for multivariate time series forecasting,'' in \emph{Proceedings of the 28th ACM SIGKDD Conference on Knowledge Discovery and Data Mining}, 2022, pp. 1567--1577.

\bibitem{liu2023fdti}
Z.~Liu, C.~Liang, G.~Zheng, and H.~Wei, ``Fdti: Fine-grained deep traffic inference with roadnet-enriched graph,'' \emph{arXiv preprint arXiv:2306.10945}, 2023.

\bibitem{wang2023pattern}
B.~Wang, Y.~Zhang, X.~Wang, P.~Wang, Z.~Zhou, L.~Bai, and Y.~Wang, ``Pattern expansion and consolidation on evolving graphs for continual traffic prediction,'' in \emph{Proceedings of the 29th ACM SIGKDD Conference on Knowledge Discovery and Data Mining}, 2023, pp. 2223--2232.

\bibitem{fan2023spatial}
Y.~Fan, C.-C.~M. Yeh, H.~Chen, Y.~Zheng, L.~Wang, J.~Wang, X.~Dai, Z.~Zhuang, and W.~Zhang, ``Spatial-temporal graph boosting networks: Enhancing spatial-temporal graph neural networks via gradient boosting,'' in \emph{Proceedings of the 32nd ACM International Conference on Information and Knowledge Management}, 2023, pp. 504--513.

\bibitem{snell2017prototypical}
J.~Snell, K.~Swersky, and R.~Zemel, ``Prototypical networks for few-shot learning,'' \emph{Advances in neural information processing systems}, vol.~30, 2017.

\bibitem{lee2022meta}
H.-y. Lee, S.-W. Li, and N.~T. Vu, ``Meta learning for natural language processing: A survey,'' \emph{arXiv preprint arXiv:2205.01500}, 2022.

\bibitem{finn2017model}
C.~Finn, P.~Abbeel, and S.~Levine, ``Model-agnostic meta-learning for fast adaptation of deep networks,'' in \emph{International conference on machine learning}.\hskip 1em plus 0.5em minus 0.4em\relax PMLR, 2017, pp. 1126--1135.

\bibitem{wei2016transfer}
Y.~Wei, Y.~Zheng, and Q.~Yang, ``Transfer knowledge between cities,'' in \emph{Proceedings of the 22nd ACM SIGKDD International Conference on Knowledge Discovery and Data Mining}, 2016, pp. 1905--1914.

\bibitem{zhang2022strans}
Y.~Zhang, Y.~Li, X.~Zhou, X.~Kong, and J.~Luo, ``Strans-gan: Spatially-transferable generative adversarial networks for urban traffic estimation,'' in \emph{2022 IEEE International Conference on Data Mining}.\hskip 1em plus 0.5em minus 0.4em\relax IEEE, 2022, pp. 743--752.

\bibitem{liu2023cross}
Z.~Liu, G.~Zheng, and Y.~Yu, ``Cross-city few-shot traffic forecasting via traffic pattern bank,'' in \emph{Proceedings of the 32nd ACM International Conference on Information and Knowledge Management}, 2023, pp. 1451--1460.

\bibitem{vaswani2017attention}
A.~Vaswani, N.~Shazeer, N.~Parmar, J.~Uszkoreit, L.~Jones, A.~N. Gomez, {\L}.~Kaiser, and I.~Polosukhin, ``Attention is all you need,'' \emph{Advances in neural information processing systems}, vol.~30, 2017.

\bibitem{zeng2023transformers}
A.~Zeng, M.~Chen, L.~Zhang, and Q.~Xu, ``Are transformers effective for time series forecasting?'' in \emph{Proceedings of the AAAI conference on artificial intelligence}, vol.~37, no.~9, 2023, pp. 11\,121--11\,128.

\bibitem{das2023long}
A.~Das, W.~Kong, A.~Leach, R.~Sen, and R.~Yu, ``Long-term forecasting with tide: Time-series dense encoder,'' \emph{arXiv preprint arXiv:2304.08424}, 2023.

\bibitem{rao2022fogs}
X.~Rao, H.~Wang, L.~Zhang, J.~Li, S.~Shang, and P.~Han, ``Fogs: First-order gradient supervision with learning-based graph for traffic flow forecasting,'' in \emph{Proceedings of International Joint Conference on Artificial Intelligence, IJCAI}.\hskip 1em plus 0.5em minus 0.4em\relax ijcai. org, 2022.

\bibitem{zhou2022fedformer}
T.~Zhou, Z.~Ma, Q.~Wen, X.~Wang, L.~Sun, and R.~Jin, ``Fedformer: Frequency enhanced decomposed transformer for long-term series forecasting,'' in \emph{International Conference on Machine Learning}.\hskip 1em plus 0.5em minus 0.4em\relax PMLR, 2022, pp. 27\,268--27\,286.

\bibitem{du2021adarnn}
Y.~Du, J.~Wang, W.~Feng, S.~Pan, T.~Qin, R.~Xu, and C.~Wang, ``Adarnn: Adaptive learning and forecasting of time series,'' in \emph{Proceedings of the 30th ACM International Conference on Information \& Knowledge Management}, 2021, pp. 402--411.

\bibitem{rousseeuw1987silhouettes}
P.~J. Rousseeuw, ``Silhouettes: a graphical aid to the interpretation and validation of cluster analysis,'' \emph{Journal of computational and applied mathematics}, vol.~20, pp. 53--65, 1987.

\end{thebibliography}


 

\begin{IEEEbiography}[{\includegraphics[width=1in,height=1.25in,clip,keepaspectratio]{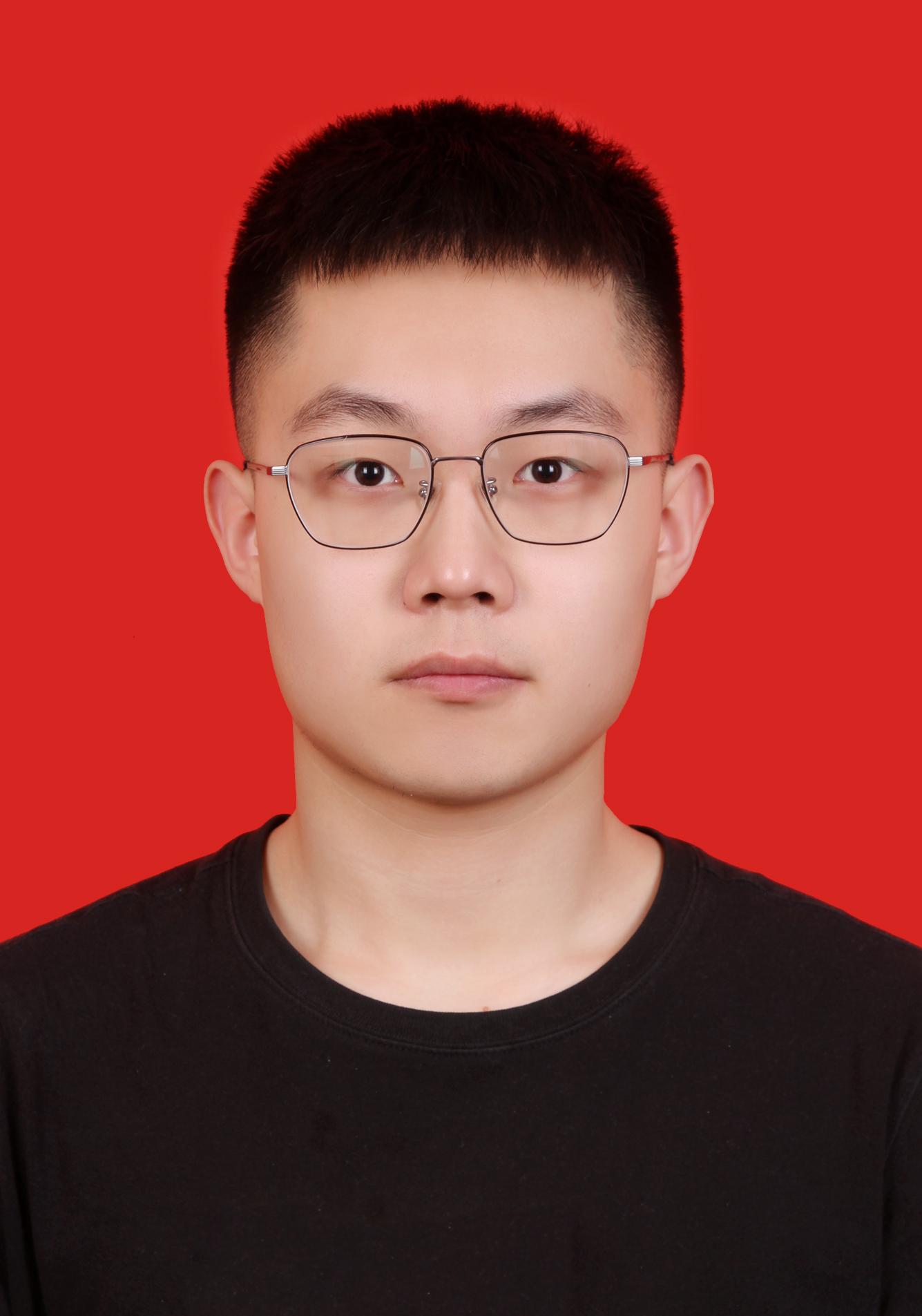}}]{Zhanyu Liu}
Received the B.E. degree from University of Electronic Science and Technology of China, Chengdu, China, in 2021. He is currently pursuing a Ph.D. degree in Computer Science at Shanghai Jiao Tong University, China. His research interests include spatial-temporal data mining, time series analysis, and data distillation. He has published several papers in top conferences, such as SIGKDD, CIKM, ECMLPKDD, etc.
\end{IEEEbiography}

\begin{IEEEbiography}[{\includegraphics[width=1in,height=1.25in,clip,keepaspectratio]{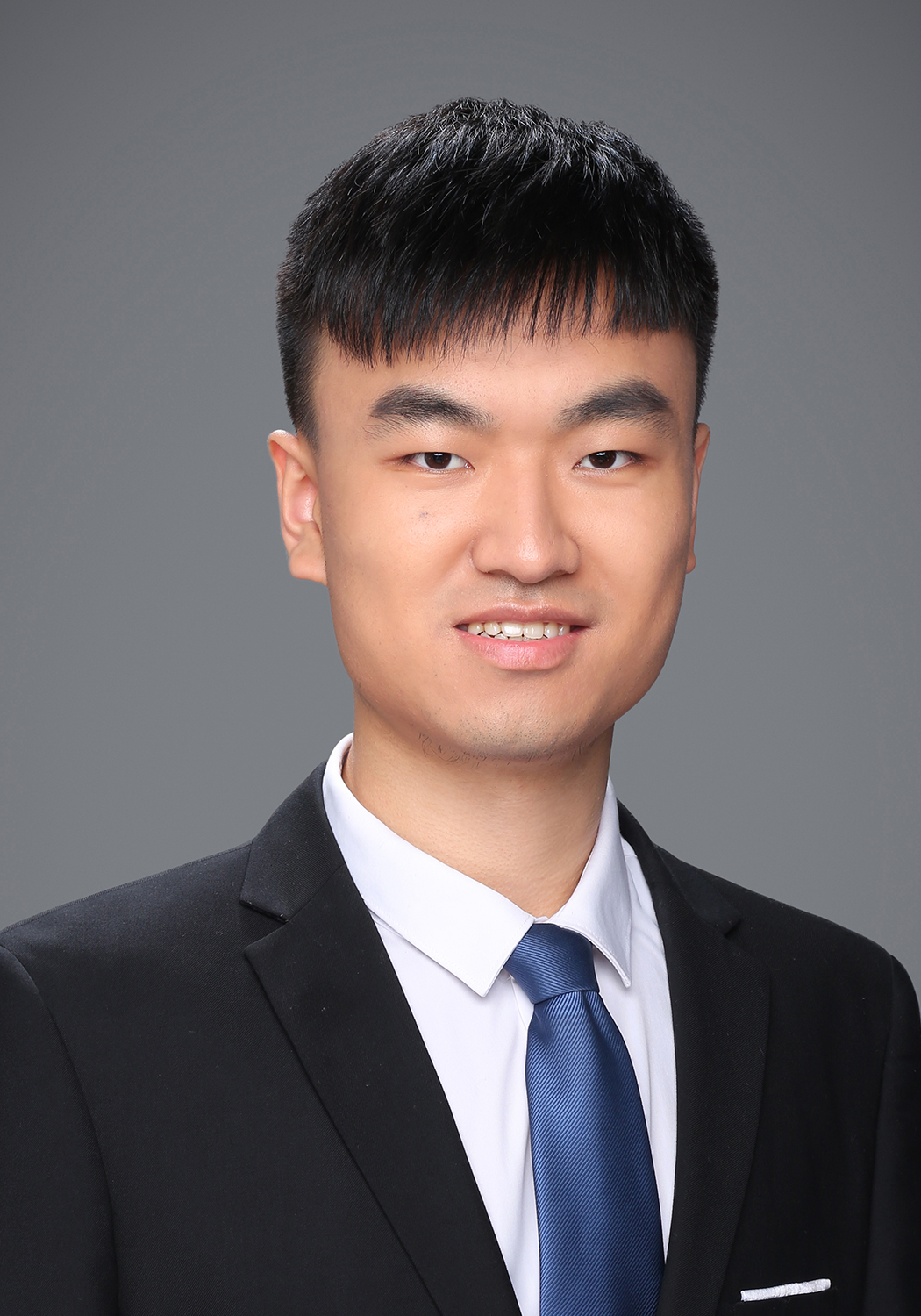}}]{Guanjie Zheng}
is an assistant professor at the John Hopcroft Center, Shanghai Jiao Tong University. He gets his Ph. D. from The Pennsylvania State University in 2020. His research interests lie in urban computing, city brain and spatio-temporal data mining. His recent work focuses on how to learn optimal strategies for city-level traffic and mobility coordination from multi-modal data. He has published more than 30 papers on top-tier conferences and journals, such as KDD, WWW, AAAI, ICDE, and CIKM. He has been cited for more than 3800 times.
\end{IEEEbiography}

\begin{IEEEbiography}[{\includegraphics[width=1in,height=1.25in,clip,keepaspectratio]{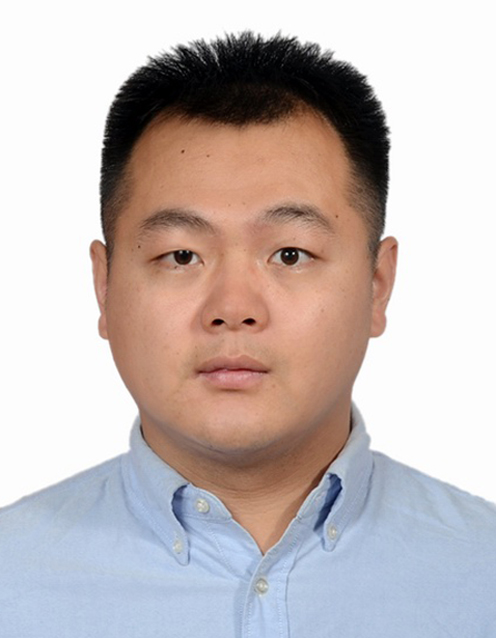}}]{Yanwei Yu} is currently a professor in the College of Computer Science and Technology of Ocean University of China. He received the B.S. degree from Liaocheng University, China, in 2008 and the Ph.D. degree from University of Science and Technology Beijing, China, in 2014, respectively. From 2012 to 2013, he was a visiting Ph.D. student at the Department of Computer Science of Worcester Polytechnic Institute. From 2016 to 2018, he was a postdoc researcher at the College of Information Sciences and Technology of Pennsylvania State University. His research interests include data mining, machine learning, and database systems. He is a member of IEEE and ACM.
\end{IEEEbiography}



\end{document}